\documentclass[]{hf-wm}

\usepackage[utf8]{inputenc}
\usepackage[T1]{fontenc}
\usepackage{lmodern}
\usepackage{url}
\usepackage{booktabs}
\usepackage{amsfonts}
\usepackage{amsmath}
\usepackage{amssymb}
\usepackage{microtype}
\microtypesetup{expansion=false}
\usepackage{xcolor}
\usepackage{textcomp}
\usepackage{graphicx}
\usepackage{array}
\usepackage{multirow}
\usepackage{xspace}
\usepackage{enumitem}
\usepackage{wrapfig}

\DeclareCaptionFormat{hfwmjustified}{{\sffamily\bfseries #1#2}#3\par}
\definecolor{gripOpen}{HTML}{246EA0}
\definecolor{gripClosed}{HTML}{B35D16}
\definecolor{gripUncertain}{HTML}{626B75}
\newcommand{\gripopen}{\textcolor{gripOpen}{Open}}
\newcommand{\gripclosed}{\textcolor{gripClosed}{Closed}}
\newcommand{\gripuncertain}{\textcolor{gripUncertain}{Indeterminate}}
\newlength{\gripTableWidth}

\newcolumntype{G}[1]{%
    >{\begin{minipage}[c][8.5mm][c]{#1}%
      \raggedright\arraybackslash\strut}%
    l%
    <{\strut\end{minipage}}%
}

\newcolumntype{H}[1]{%
    >{\begin{minipage}[c][8.5mm][c]{#1}%
      \centering\arraybackslash\strut}%
    l%
    <{\strut\end{minipage}}%
}

\newcommand{\method}{HF-WM}
\providecommand{\keywords}[1]{\noindent\textbf{Keywords:} #1}

\title{{\fontsize{14.5}{18}\selectfont\textbf{Beyond Successful Robot Demonstrations: Learning from\\\textcolor{wamfont}{H}uman Interactions and Modeling \textcolor{wamfont}{F}ailure Dynamics}}}

\author[1,2]{Peiyan Tu}
\author[5,2]{Cong Wang}
\author[3]{Hanxin Zhu}
\author[4,2\ddagger]{Xin Jin}
\author[3,2,\ddagger]{Zhibo Chen}

\affiliation[1]{Zhejiang University}
\affiliation[2]{Beijing Zhongguancun Academy}
\affiliation[3]{University of Science and Technology of China}
\affiliation[4]{Eastern Institute of Technology}
\affiliation[5]{Institute of Automation, Chinese Academy of Sciences}
\contribution[\dagger]{Project Leader}
\contribution[\ddagger]{Equal Corresponding Author}

\abstract{
    Action-conditioned world models offer a promising foundation for generalist robotic intelligence by predicting the consequences of actions.
    However, existing robotic datasets provide limited coverage of the diverse interaction dynamics required for generalizable world modeling. Large-scale egocentric human videos offer complementary interaction experience, yet the embodiment gap between human hands and robotic manipulators hinders effective cross-embodiment learning. Meanwhile, the predominance of successful demonstrations leaves failure dynamics substantially underrepresented, limiting the ability of world models to anticipate unsuccessful executions.
    To address these limitations, we introduce \textbf{HF-WM}, an action-conditioned world model that combines \textbf{H}uman interaction learning with \textbf{F}ailure-aware post-training. To bridge the embodiment gap, we develop a dual-branch framework that couples video generation with auxiliary pointflow prediction, using stabilized pointflow as a unified motion representation across human and robot interactions while suppressing egocentric camera motion. To improve failure modeling, we further perform Group Relative Policy Optimization (GRPO) post-training on perturbed actions derived from successful demonstrations, guided by action alignment and physical plausibility rewards, enabling the model to learn plausible failure dynamics without paired ground-truth videos.
    Extensive experiments demonstrate that HF-WM achieves state-of-the-art performance in action-conditioned video generation and exhibits strong utility for policy evaluation in downstream robotic tasks.

}

\date{\today}
\project{\url{https://peiyantu.github.io/HF-WM/}}

\begin{document}
\thispagestyle{firstheader}
\maketitle


\begin{figure}[!b]
    \centering
    \includegraphics[width=\linewidth]{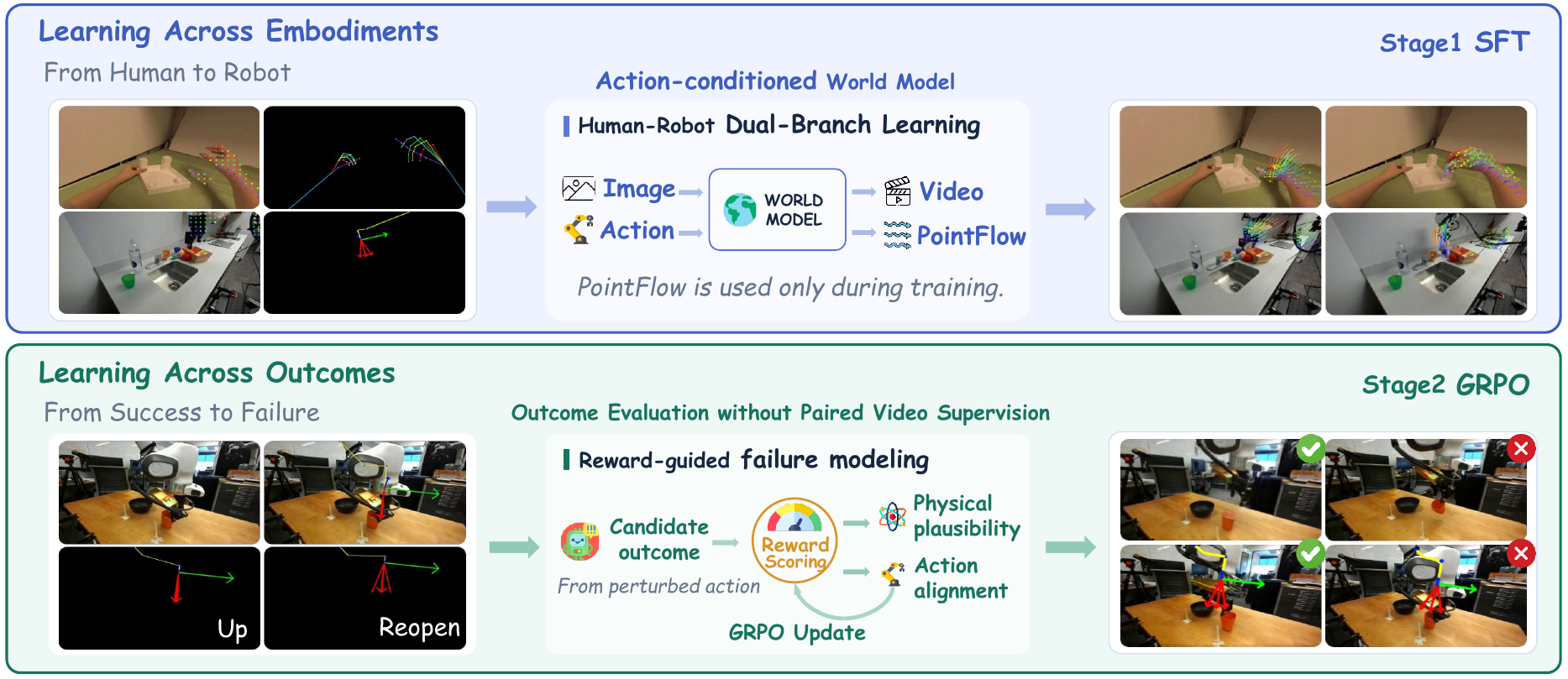}
    \caption{\textbf{Overview of HF-WM architecture.} HF-WM extends learning beyond successful robot demonstrations in two stages. Stage 1 uses a dual-branch architecture with stabilized pointflow supervision to learn interaction dynamics from egocentric human interactions (Sec.~\ref{human}). Stage 2 applies GRPO to counterfactual action perturbations, guided by action alignment and physical plausibility rewards, to model failure dynamics without paired failure-video supervision (Sec.~\ref{grpo}).}
    \label{fig:teaser}
\end{figure}

\section{Introduction}
With the rapid advancement of video generation models \citep{wan, cosmos3}, world models have witnessed significant breakthroughs and are increasingly regarded as a promising pathway toward generalist robotics \citep{dreamzero, fastwam}. In particular, action-conditioned world models take robot action sequences as input and model the resulting state transitions to predict future states \citep{irasim}. This predictive capability supports scalable data generation, policy evaluation, and reinforcement learning in generated environments \citep{EVA, pilworld, sc3}, thereby closing the loop between imagination, evaluation, and policy improvement. This paradigm has emerged as a key foundation for learning and improving robot action planning systems.

Despite these advances, action-conditioned world models still struggle to accurately capture interaction dynamics and execution outcomes, largely due to the limited scale and coverage of high-quality embodied data~\citep{videoembody}.
To alleviate this limitation, recent studies have turned to large scale egocentric human object interaction datasets \citep{dreamdojo,oscar}, leveraging human demonstrations to enhance model performance. \textbf{1)} However, a substantial domain gap persists between human and robotic manipulation, and egocentric recordings are often affected by camera jitter and unstable viewpoints, which can obscure interaction-relevant regions and hinder the learning of physically meaningful representations. Consequently, effectively exploiting human egocentric data for embodied world model learning remains an open challenge \citep{videoembody}.
\textbf{2)} In addition, existing embodied datasets are heavily biased toward successful demonstrations, with limited coverage of failure cases and off-target interactions \citep{mirabench}. Yet world models for embodied scenarios require not only modeling successful trajectories, but also capturing failure dynamics and counterfactual outcomes induced by erroneous or suboptimal actions \citep{pilworld}. In the absence of failure demonstrations, how to effectively synthesize failure scenarios to improve physical reasoning remains an important open problem.

\section{Related Work}

To this end, as shown in Fig.~\ref{fig:teaser}, we propose a set of key insights to address the aforementioned challenges. \textbf{1)} Firstly, we introduce a skeleton grounded action representation that bridges human and robotic action spaces through a shared spatial formulation \citep{oscar}, mitigating the cross embodiment gap. To better exploit egocentric human interaction data, we introduce pointflow supervision as a motion constraint that tracks point trajectories, capturing interaction driven dynamics \citep{pointworld, tracegen}. We further apply a stabilization procedure on pointflows to suppress camera induced motion, ensuring that point trajectories primarily reflect interaction driven dynamics rather than viewpoint variations. This allows the model to better focus on physically meaningful motion signals and improves the effective utilization of  egocentric human interaction data. The primary challenge in introducing failure cases is the lack of real failure video supervision, which limits direct learning under failure scenarios.
\textbf{2)} To address this, we propose a Group Relative Policy Optimization (GRPO) based reinforcement learning framework \citep{flowgrpo} for post-training, enabling better modeling of failure dynamics using reward signals without requiring ground-truth failure video supervision.
Specifically, we construct pseudo failure actions by perturbing successful action trajectories, thereby inducing counterfactual action conditions. These perturbed actions are then used as conditions for video generation, where the model is optimized via reward-based feedback under GRPO.
The reward function consists of two complementary terms: a skeleton-based action following score that measures the consistency between the generated motion and the given action condition \citep{abot}, and a physical plausibility score that evaluates whether the predicted dynamics remain physically consistent. Both rewards are computable without paired ground-truth videos, enabling GRPO post-training on perturbed actions to strengthen failure-aware dynamics modeling without direct supervision of the corresponding outcomes.

In summary, our main contributions are:
\begin{itemize}[leftmargin=*, nosep]
    \item We present HF-WM, an action-conditioned world model that learns from egocentric human interactions and counterfactual action perturbations for embodied dynamics modeling.

    \item We introduce a dual-branch framework with auxiliary pointflow supervision, establishing a shared motion representation across human and robot interactions that enables embodied world models to learn more effectively from egocentric human data.

    \item We present a GRPO based post-training framework that enables failure aware optimization without paired failure video supervision.

    \item {\sloppy Extensive experiments demonstrate that HF-WM achieves state-of-the-art performance in action-conditioned video generation, and further validate its effectiveness as a policy evaluator for downstream robotic tasks.\par}
\end{itemize}

\noindent\textbf{Controllable video generation.}
Controllable video generation as an important extension of video foundation models, aims to introduce diverse external control signals beyond text conditioning to enable fine-grained control over generated content. Recent work on controllable generation incorporates a variety of conditioning signals into existing video foundation models, including spatial structural information (e.g., human poses \citep{egosim}, depth maps \citep{gta}, and skeletons \citep{oscar}), temporal dynamics (e.g., optical flow \citep{orthophys}, motion trajectories \citep{irasim}, and camera parameters \citep{tu2026embody4d}), as well as multimodal inputs such as reference images or audio. These methods typically inject external control signals into pretrained video diffusion models through dedicated condition encoders \citep{tesseract}, cross-attention mechanisms \citep{trajectorycrafter}, or ControlNet architectures \citep{yang2026neoverse}, thereby constraining and guiding the generation process.

\noindent\textbf{Action-conditioned world models.}
Recent advances have positioned action-conditioned world models as a key interface between generative video modeling, action-conditioned dynamics prediction, and policy evaluation in embodied intelligence \citep{abot}. IRASim \citep{irasim} incorporates robotic trajectories into the video generation process, demonstrating the feasibility of synthesizing future visual states under explicit action conditions. Ctrl-World \citep{ctrl} extends this paradigm to multi-view embodied generation, where three-view synthesis improves the spatial coverage of generated environments and yields practical gains in real-robot deployment. SC3-Eval \citep{sc3} further closes the loop by leveraging world models as offline policy evaluators, enabling policies to be assessed without costly physical rollouts.
Despite recent progress, existing action-conditioned world models remain constrained by limited embodied data and failure coverage \citep{dreamdojo, oscar}, motivating us to study how human object interaction data and counterfactual failure samples can improve the modeling of action consequences and physical dynamics.

\section{Method}
\label{sec:method}

\subsection{Preliminaries}
\textbf{Mixture-of-Transformers Foundation Model.}
We build on Cosmos~3~\citep{cosmos3}, an omnimodal world model that jointly
models language, video, and actions using a Mixture-of-Transformers (MOT)
architecture. It factorizes multimodal sequences into an autoregressive stream
for text and ViT encoded visual tokens, and a diffusion stream for
VAE encoded video latents and action tokens. Shared multimodal attention
couples the two streams: autoregressive tokens use causal attention, whereas
diffusion tokens attend bidirectionally to the autoregressive context and
diffusion states.
For visual generation, the Wan2.2 VAE~\citep{wan} encodes videos into
spatio-temporal latents. Given a clean latent
$\mathbf{x}_0$, Gaussian noise
$\boldsymbol{\epsilon}\sim\mathcal{N}(\mathbf{0},\mathbf{I})$, and a flow time
$t\in[0,1]$, the interpolated latent is
\begin{equation}
    \mathbf{x}_t
    =(1-t)\mathbf{x}_0+t\boldsymbol{\epsilon}.
\end{equation}
The generator is trained with the rectified-flow matching objective
\begin{equation}
    \mathcal{L}_{\mathrm{RF}}
    =
    \mathbb{E}_{\mathbf{c},t,\mathbf{x}_0,\boldsymbol{\epsilon}}
    \left[
        \left\|
        v_{\theta}(\mathbf{x}_t,t,\mathbf{c})
        -(\boldsymbol{\epsilon}-\mathbf{x}_0)
        \right\|_2^2
    \right],
\end{equation}
where $v_{\theta}$ is the predicted velocity field and $\mathbf{c}$ denotes
the conditioning context, including text, visual observations, and action
tokens. In our setting, clean action tokens condition the denoising of future
visual latents, enabling prediction of the resulting state transitions.

\textbf{Flow-Based Group Relative Policy Optimization.}
We formulate reward-guided rectified-flow denoising as a sequential decision
process. Given a conditioning input $\mathbf{c}$, the policy
$\pi_{\theta}$ is induced by the velocity field
$v_{\theta}(\mathbf{x}_t,t,\mathbf{c})$. Following Flow-GRPO~\citep{flowgrpo},
we replace the deterministic rectified-flow ODE with the marginal-preserving
SDE
\begin{equation}
\begin{aligned}
d\mathbf{x}_t
={}&\left[
v_{\theta}(\mathbf{x}_t,t,\mathbf{c})
+\frac{\sigma_t^2}{2t}
\left(
\mathbf{x}_t
+(1-t)v_{\theta}(\mathbf{x}_t,t,\mathbf{c})
\right)
\right]dt
+\sigma_t\,d\mathbf{w}_t,
\qquad t\in(0,1],
\end{aligned}
\end{equation}
where $\sigma_t$ controls the exploration strength and
$d\mathbf{w}_t$ is a Wiener increment.

After discretizing the denoising process into $N$ steps, we sample $K$
trajectories
$\tau^k=(\mathbf{x}_N^k,\ldots,\mathbf{x}_0^k)$ from the old policy. At each
step $n=1,\ldots,N$, the transition ratio and its clipped version are
\begin{equation}
    \rho_n^k(\theta)
    =
    \frac{
    \pi_{\theta}(\mathbf{x}_{n-1}^k\mid\mathbf{x}_n^k,\mathbf{c})
    }{
    \pi_{\theta_{\mathrm{old}}}(\mathbf{x}_{n-1}^k\mid\mathbf{x}_n^k,\mathbf{c})
    },
    \qquad
    \bar{\rho}_n^k(\theta)
    =
    \operatorname{clip}\!\left(
    \rho_n^k(\theta),1-\varepsilon,1+\varepsilon
    \right).
\end{equation}

The terminal reward is normalized within each group:
\begin{equation}
    A^k
    =
    \frac{
    r(\mathbf{x}_0^k,\mathbf{c})-\mu_r
    }{
    \sigma_r+\delta
    },
    \qquad
    (\mu_r,\sigma_r)
    =
    \operatorname{MeanStd}_{j=1}^{K}
    \left\{
    r(\mathbf{x}_0^j,\mathbf{c})
    \right\},
\end{equation}
where $\delta>0$ is a small numerical stabilizer. The resulting GRPO
objective is
\begin{equation}
\mathcal{J}(\theta)=\frac{1}{K}\sum_{k=1}^{K}\frac{1}{N}\sum_{n=1}^{N}\left[\min\!\left(\rho_n^k(\theta)A^k,\bar{\rho}_n^k(\theta)A^k\right)-\beta D_{\mathrm{KL}}\!\left(\pi_{\theta}(\cdot\mid\mathbf{x}_n^k,\mathbf{c})\,\middle\|\,\pi_{\mathrm{ref}}(\cdot\mid\mathbf{x}_n^k,\mathbf{c})\right)\right].
\end{equation}
This formulation allows the generator to be optimized with pixel-level rewards from generated samples, without paired video ground-truth; group-relative feedback further improves generation quality and alignment.

\begin{figure}[!t]
    \centering
    \includegraphics[width=1.0\linewidth]{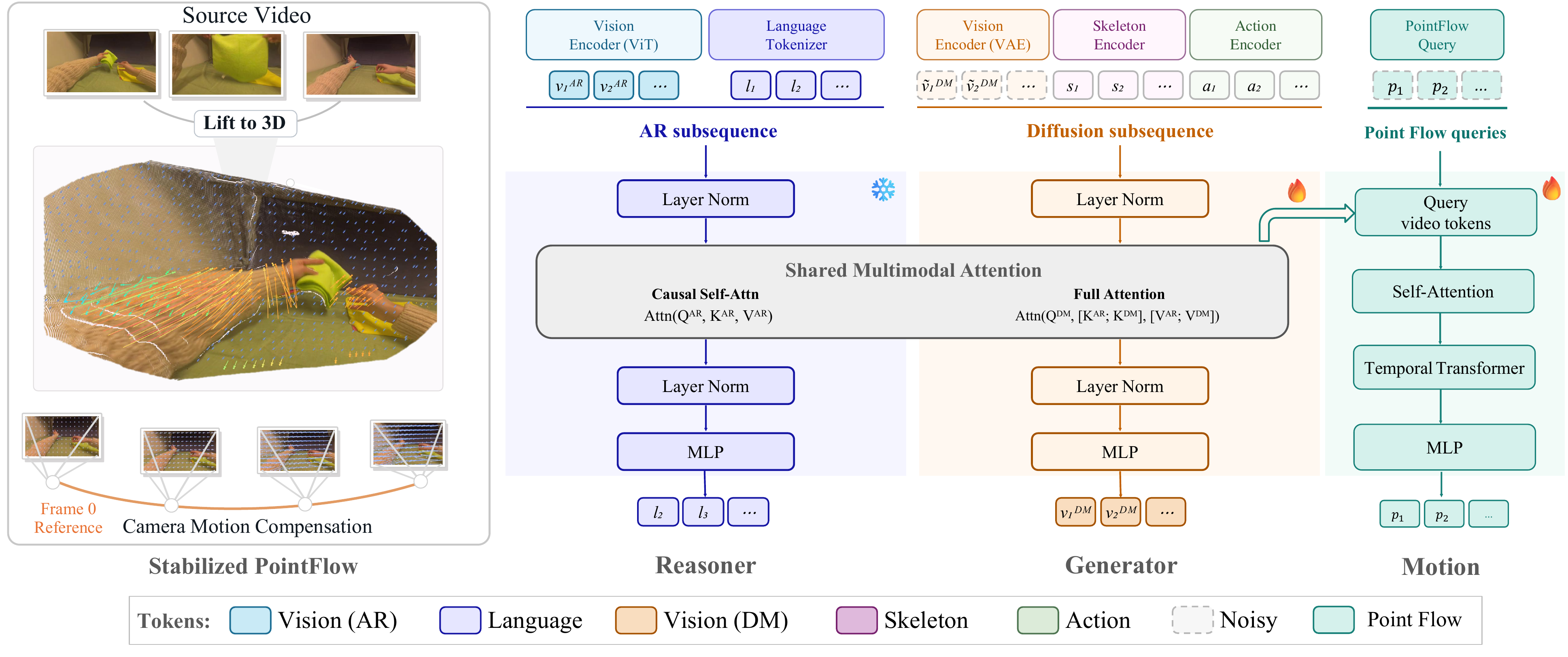}
    \caption{\textbf{Dual-branch architecture of \method{} with stabilized pointflow.} The auxiliary pointflow branch predicts interaction motion from intermediate video representations, while pointflow stabilization suppresses egocentric camera motion so that supervision is better focused on interaction-induced dynamics. The pointflow branch is used only during training and introduces no additional inference overhead.}
    \label{fig:pointflow}
\end{figure}
\vspace{-0.3cm}
\subsection{Learning Across Embodiments: From Human to Robot}
\label{human}
Introducing human egocentric interaction data into action-conditioned world models first requires addressing the representational mismatch between human hand actions and robot arm actions. These two action forms differ naturally in degrees of freedom, dimensionality, and motion structure, making it difficult to share a common action conditioning space directly. Thanks to the Cosmos 3 foundation model \citep{cosmos3}, which has been pretrained on both human interaction data and embodied manipulation data, the model already possesses certain cross embodiment interaction priors, thereby partially reducing the domain gap between human hands and robot arms. Building on this prior, we adopt the 2D skeleton representation from OSCAR \citep{oscar}, mapping human hand and robot arm actions into a shared spatial action space, thereby avoiding direct alignment between heterogeneous action vectors. With this unified action representation, human egocentric interaction data can be incorporated as effective augmentation data for training action-conditioned world models. Furthermore, we develop a dual-branch learning framework that couples video generation with auxiliary pointflow prediction \citep{pointworld}, where pointflow supervision guides the video branch to focus on motion regions that are truly relevant to physical interactions, enabling human interaction data to provide more effective supervision for embodied world model learning.

For the auxiliary pointflow branch, we introduce a stabilized pointflow representation to mitigate perturbations caused by egocentric camera motion in human object interaction data, thereby encouraging the model to focus on interaction induced dynamics. We first use CoWTracker \citep{cowtracker} to obtain 2D point trajectories $\mathbf{p}_{i,t}=(u_{i,t},v_{i,t})$, where $i$ and $t$ index points and frames, and $(u,v)$ denote image coordinates.
To disentangle camera motion from the tracked trajectories, we use monocular depth estimated with Depth Anything 3 \citep{lin2025depth} together with the camera parameters provided by the dataset. For each tracked point $\mathbf{p}_{i,0}$, we back-project its first frame location into 3D as
\begin{equation}
    \mathbf{X}_{i,0}
    =
    D_0(\mathbf{p}_{i,0})
    \mathbf{K}_0^{-1}
    \tilde{\mathbf{p}}_{i,0},
    \qquad
    \tilde{\mathbf{p}}_{i,0}
    =
    [u_{i,0},v_{i,0},1]^{\mathsf T},
\end{equation}
where $D_0$ and $\mathbf{K}_0$ denote the depth map and camera
intrinsics of the first frame, respectively. We then transform $\mathbf{X}_{i,0}$ to frame $t$ using the corresponding camera extrinsics and reproject it onto the image plane:
\begin{equation}
    \hat{\mathbf{p}}^{\mathrm{cam}}_{i,t}
    =
    \pi\!\left(
        \mathbf{K}_t
        \mathcal{T}_{0\rightarrow t}(\mathbf{X}_{i,0})
    \right),
    \qquad
    \mathbf{f}^{\mathrm{stab}}_{i,t}
    =
    \mathbf{p}_{i,t}
    -
    \hat{\mathbf{p}}^{\mathrm{cam}}_{i,t},
\end{equation}
where $\mathcal{T}_{0\rightarrow t}$ maps 3D points from the camera
coordinate system of frame $0$ to that of frame $t$,
$\pi$ denotes perspective division, and $\mathbf{f}^{\mathrm{stab}}_{i,t}$ denotes the
camera-motion-compensated displacement.
By suppressing motion caused by viewpoint changes, the stabilized
pointflow enables the model to better learn interaction dynamics
from human demonstration data.

As shown in Fig.~\ref{fig:pointflow}, our dual-branch architecture consists of a video generation branch and a lightweight auxiliary pointflow branch, which provides shared motion supervision across human and robot interactions. It reads intermediate video latents from layers 4, 6, 8, and 10, and predicts the flow matching velocity of noisy future pointflow using the first frame visual latent and initial point coordinates as additional inputs.
Pointflow tokens are updated through self attention and cross attention to the intermediate video representations. Importantly, the pointflow branch attends only to the intermediate representations of the generated video, without direct access to the action sketch conditioning tokens. This design avoids over reliance on the action conditioning signal and encourages pointflow to capture motion from the generated visual dynamics, providing more effective supervision for the video backbone.

We train the pointflow branch with the following flow-matching objective:
\begin{equation}
\mathcal{L}_{\mathrm{point}}
=
\mathbb{E}_{\rho,\boldsymbol{\epsilon}}
\left[
\left\|
g_{\phi}\!\left(
\mathbf{S}_{\rho},
\rho,
\mathbf{z}_0,
\mathbf{p}_0;
\mathbf{H}_{\theta}
\right)
-
(\boldsymbol{\epsilon}-\mathbf{S})
\right\|_2^2
\right],
\quad
\mathbf{S}_{\rho}
=
(1-\rho)\mathbf{S}
+
\rho\boldsymbol{\epsilon},
\end{equation}
where \(\mathbf{S}\) denotes the pointflow supervision target formed
by stacking \(\mathbf{f}^{\mathrm{stab}}_{i,t}\) over tracked points
and future frames,
\(\mathbf{S}_{\rho}\) is its noisy interpolation at noise level
\(\rho\in[0,1]\),
and \(\boldsymbol{\epsilon}\) is Gaussian noise.
\(\mathbf{z}_0\) and \(\mathbf{p}_0\) denote the first-frame visual latent
and point coordinates, respectively.
\(\mathbf{H}_{\theta}\) represents the intermediate video
latents extracted from the four selected backbone layers, and
\(g_{\phi}\) denotes the pointflow head.
The two branches are jointly optimized with the original video flow-matching objective:
\begin{equation}
\mathcal{L}
=
\mathcal{L}_{\mathrm{video}}
+
\lambda_p\mathcal{L}_{\mathrm{point}},
\end{equation}
where \(\mathcal{L}_{\mathrm{video}}\) is the video flow-matching loss
\(\mathcal{L}_{\mathrm{RF}}\) defined in the preliminaries, and
\(\lambda_p\) controls the contribution of pointflow supervision.
The auxiliary pointflow branch is used only during training and does not alter the video generation pathway. Its gradients propagate through the intermediate video latents to the backbone, encouraging representations that better capture interaction-induced motion, while introducing no additional overhead at inference time.

\subsection{Learning Across Outcomes: From Success to Failure}
\label{grpo}

Robot manipulation datasets are dominated by successful demonstrations, leaving failure dynamics substantially underrepresented. Consequently, supervised fine-tuning on such data provides limited supervision for the consequences of abnormal gripper operations, mistimed actuation, and deviations in end-effector pose. To overcome this limitation, we introduce a second-stage GRPO post-training framework that learns failure dynamics from counterfactual action perturbations.

We first classify each action segment according to its gripper-state transition pattern and select the perturbation accordingly. For trajectories with clear gripper transitions, we perturb either the gripper state and transition timing or the end-effector pose, covering different failure modes that may arise during interaction. For trajectories without clear gripper events, we primarily perturb the end-effector pose through rotation and translation, focusing optimization on action following and motion consistency while avoiding gripper changes inconsistent with the original interaction.

The counterfactual actions do not have corresponding ground-truth future videos, making direct optimization with supervised video losses infeasible. To address this, we apply GRPO by generating multiple candidate futures under the same counterfactual action condition and performing group-relative optimization with action alignment and physical plausibility rewards. The action alignment reward measures the consistency between generated motion and the conditioning action, while the physical plausibility reward evaluates the physical validity of contact, object motion, and robot structure.

\textbf{Action Alignment Reward.}
We train a visual action alignment scorer with pairwise supervision.
Each training comparison consists of two video--action samples,
$(V,A^{+})$ and $(V,A^{-})$, sharing the same RGB video but containing
different action skeletons. Comparisons include original actions
against perturbed variants, as well as variants with different
perturbation strengths, covering pose offsets, temporal shifts,
and gripper-state changes.
We overlay its action skeleton on the corresponding RGB frames,
using the full-frame sequence as the global input and crops around
the end effector as the local input. The global branch encodes
overall robot pose and motion, while the local branch focuses on
the wrist, gripper, and interaction region. Their features are
fused and aggregated by a temporal Transformer. The two samples
are evaluated independently by the same scoring network, producing
predictions $z^{+}$ and $z^{-}$. We convert these predictions into
a temperature-scaled preference probability and a signed
preference gap:
\begin{equation}
p
=
\sigma\!\left(\frac{z^{+}-z^{-}}{\tau}\right),
\qquad
r_{\Delta}
=
2p-1,
\label{eq:align_gap}
\end{equation}
where $p$ is the predicted probability of preferring the first
sample over the second, $\sigma$ is the sigmoid function, and
$\tau$ is a temperature parameter.
The core training objective supervises both the preference
direction and the gap magnitude:
\begin{equation}
\mathcal{L}_{\mathrm{align}}
=
\mathcal{L}_{\mathrm{rank}}
+
\lambda_{\mathrm{gap}}\mathcal{L}_{\mathrm{gap}},
\qquad
\mathcal{L}_{\mathrm{gap}}
=
\operatorname{SmoothL1}(r_{\Delta},g^{*}).
\label{eq:align_loss}
\end{equation}
The ranking term encourages a higher score for the better-aligned
sample, while the gap term matches the predicted preference gap
to a target $g^{*}$ specified by the perturbation type, relative
severity, and visual discernibility. Subtle differences receive
smaller target gaps, whereas pronounced differences in pose,
timing, or gripper state receive larger ones. Gradients from the
pairwise objective propagate through both predictions to update
the shared scoring network.

During GRPO, the learned scorer independently evaluates candidate
rollouts conditioned on the same action sequence. Their predictions
are converted into pairwise preferences using
Eq.~\ref{eq:align_gap}, and the resulting preference gaps are used
to construct the action alignment reward.

\textbf{Physical Plausibility Reward.}
We adopt a hierarchical evaluation scheme \citep{abot} based on a shared physical checklist tailored to each counterfactual interaction. A vision-language model constructs the checklist from a multimodal prompt comprising the original demonstration video, task description, and a textual description of the action perturbation, grounding its questions in scene-specific object interactions, contact relations, and support conditions. By deriving the checklist from this context without access to candidate videos, we establish fixed evaluation criteria for all rollouts within each group.
The evaluation assesses the physical plausibility of counterfactual interactions rather than task progress, without treating task failure as a physical violation. The checklist comprises two tiers: Tier 1 identifies critical physical violations, such as object interpenetration and unsupported floating; Tier 2 provides fine-grained assessments of physical consistency, robot structure, contact causality, and visual quality. The VLM evaluates each candidate against the checklist, with judgments aggregated across dimensions to obtain the physical reward. If a candidate fails any Tier 1 check or exhibits a severe violation, we cap its physical reward and omit its action alignment reward, preventing action consistency from offsetting physical violations. Prompt templates and scoring rules are provided in the Appendix~\ref{more_grpo}.

\section{Experiments}
\label{sec:experiments}

\subsection{Experimental Setup}
We fine-tune Cosmos 3 Nano in two stages on 8 NVIDIA A100 GPUs.
The first stage performs supervised fine-tuning on robot and egocentric human interaction data with auxiliary stabilized pointflow supervision
to improve motion modeling, while the second stage applies Group Relative Policy Optimization (GRPO) to enhance the modeling of failure outcomes. All training samples adopt the unified skeleton representation provided by OSCAR \citep{oscar}. The training set contains approximately 45K Franka trajectories from RH20T \citep{rh20t}, InternData-A1 \citep{interndata}, and DROID \citep{droid}, together with 83K human interaction trajectories from EgoDex \citep{egodex} and EPIC-Kitchens \citep{epic}. Further training details are provided in the Appendix~\ref{more_ex}.

\begin{figure}[!t]
    \centering
    \includegraphics[width=1.0\linewidth]{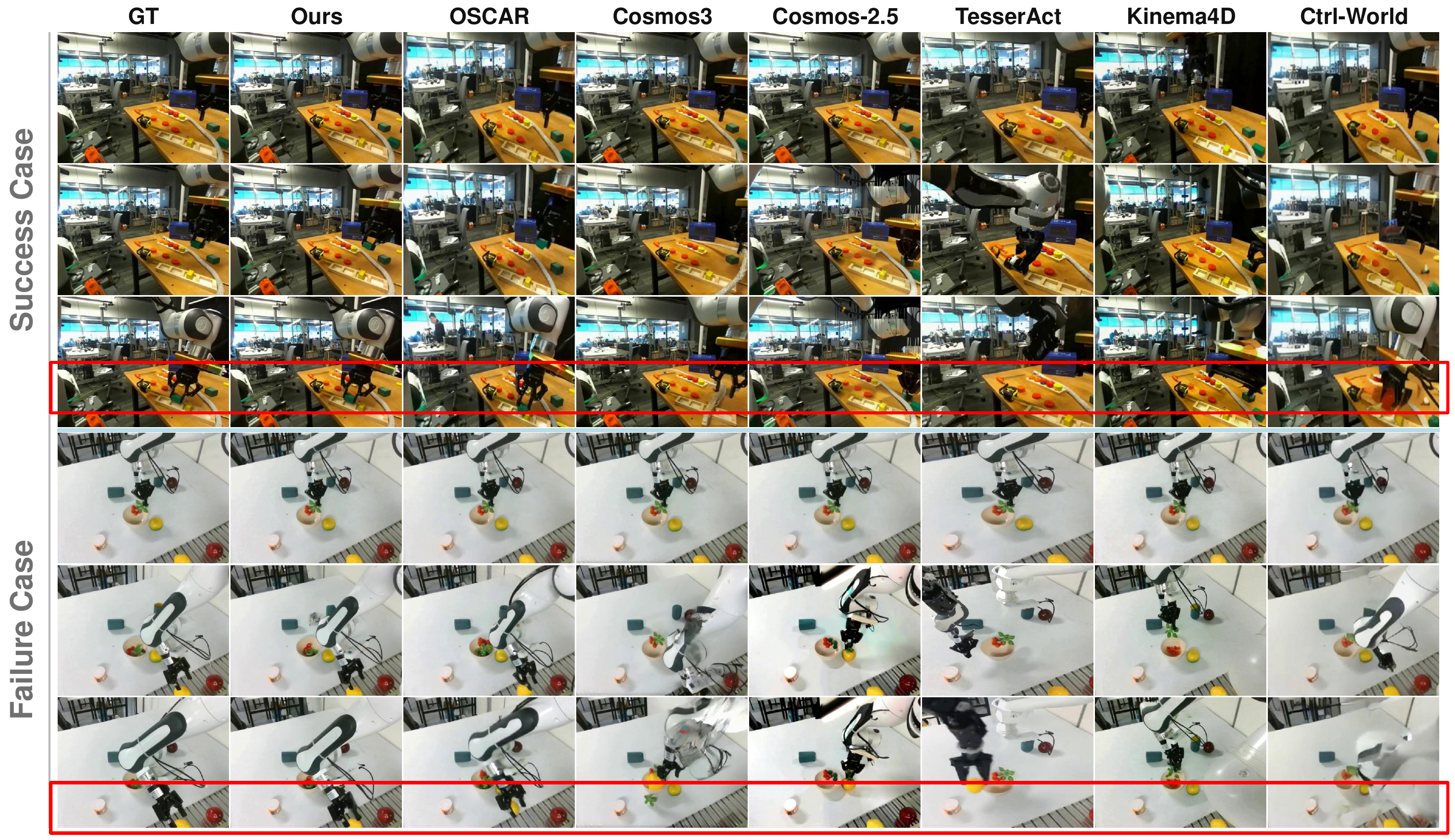}
    \caption{\textbf{Qualitative comparison of action-conditioned video generation.} Evaluated on both successful and failed execution samples, our method is compared against six models and achieves much better visual quality with precise action following.}
    \label{fig:compare}
    \vspace{-0.2cm}
\end{figure}

\begin{table*}[t]
    \centering
    \caption{Quantitative comparison with baselines. \textbf{Best} / \underline{second-best} results are bolded / underlined.}
    \label{tab:comparison}
    \resizebox{\textwidth}{!}{
    \begin{tabular}{lccccccc}
        \toprule
        \textbf{Method}
        & \textbf{PSNR}$\uparrow$
        & \textbf{SSIM}$\uparrow$
        & \textbf{LPIPS}$\downarrow$
        & \textbf{FVD}$\downarrow$
        & \textbf{FID}$\downarrow$
        & $\mathbf{L2}_{\mathrm{latent}}\downarrow$
        & \textbf{IDM}$\downarrow$ \\
        \midrule

        Cosmos-Predict2.5 \citep{cosmos}
        & 18.874 & 0.746 & 0.266
        & 60.119 & 54.108 & 0.486 & 1.268 \\

        TesserAct \citep{tesseract}
        & 17.383 & 0.757 & 0.263
        & 60.563 & 59.889 & 0.523 & 1.663 \\

        Ctrl-World \citep{ctrl}
        & 21.683 & 0.770 & 0.305
        & 36.916 & 64.665 & 0.418 & 0.949 \\

        Cosmos 3 \citep{cosmos3}
        & 20.233 & 0.766 & 0.178
        & 42.987 & 36.780 & 0.365 & 1.108 \\

        OSCAR \citep{oscar}
        & \underline{23.357}
        & \underline{0.840}
        & \underline{0.142}
        & 31.815
        & \underline{28.241}
        & \underline{0.292}
        & \underline{0.579} \\

        Kinema4D \citep{kinema4d}
        & 20.015 & 0.804 & 0.157
        & \underline{30.511} & 30.890 & 0.326 & 0.914 \\

        \midrule

        \method{} (\textbf{Ours})
        & \textbf{26.077}
        & \textbf{0.889}
        & \textbf{0.082}
        & \textbf{14.449}
        & \textbf{14.291}
        & \textbf{0.193}
        & \textbf{0.332} \\

        \bottomrule
    \end{tabular}
    }
\end{table*}

\subsection{Comparison with Baselines}
We evaluate all methods on 100 held-out Franka trajectories selected from DROID and 100 failed trajectories collected from real-world robot executions, with the reported results averaged over all evaluation trajectories. Based on how the conditioning signal is incorporated, the baselines are grouped into three categories: text-only methods (TesserAct \citep{tesseract}, Cosmos-Predict2.5 \citep{cosmos}), latent action methods (Ctrl-World \citep{ctrl} and Cosmos 3 \citep{cosmos3}), and explicit geometry methods (OSCAR \citep{oscar} and \mbox{Kinema4D~\citep{kinema4d}}).
In addition to standard visual quality metrics, we fine-tune Cosmos 3 using its inverse dynamics modeling (IDM) capability to obtain a Franka-specific IDM. It predicts actions from generated videos, which are compared with the conditioning actions to assess action following fidelity.

As shown in Fig.~\ref{fig:compare} and Table~\ref{tab:comparison}, HF-WM achieves state-of-the-art performance across visual quality, action following, and dynamics modeling, demonstrating its strong overall capability for action-conditioned world modeling.
Notably, Table~\ref{tab:comparison_success} reveals a consistent performance gap between successful and failed executions across existing models, suggesting that models trained predominantly on successful demonstrations are less effective at capturing underrepresented failure dynamics. Such an imbalance is particularly limiting for world models, which are expected to faithfully predict action consequences across diverse execution outcomes. In contrast, HF-WM maintains strong performance on both successful and failed executions, demonstrating improved robustness in modeling failure dynamics. Across both tables, HF-WM achieves the best overall performance in visual quality, outcome modeling, and action following, demonstrating more accurate and robust modeling of action-conditioned dynamics across diverse execution outcomes.

\begin{figure}[t]
    \centering
    \includegraphics[width=1.0\linewidth]{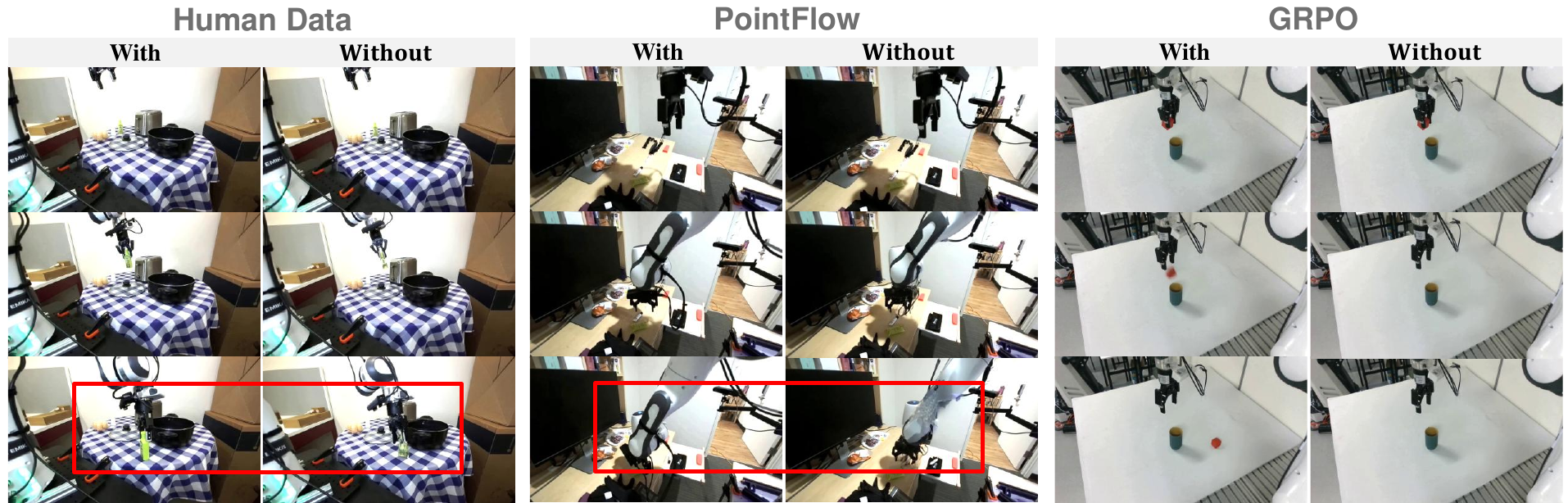}
    \caption{\textbf{Qualitative ablation of the key components in HF-WM.}
    Large-scale human interaction data broadens the diversity of object interactions, while unified pointflow supervision provides a shared motion representation that facilitates cross-embodiment transfer. Building on this foundation, GRPO post-training improves the modeling of failure dynamics across diverse execution outcomes.
    }
    \label{fig:ab}
    \vspace{-0.25cm}
\end{figure}

\begin{table*}[t]
    \centering
    \caption{Ablation study on the effectiveness of human interaction data and pointflow constraints.}
    \label{tab:human_ab}
    \resizebox{\textwidth}{!}{
    \begin{tabular}{lccccccc}
        \toprule
        \textbf{Variant}
        & \textbf{PSNR}$\uparrow$
        & \textbf{SSIM}$\uparrow$
        & \textbf{LPIPS}$\downarrow$
        & \textbf{FVD}$\downarrow$
        & \textbf{FID}$\downarrow$
        & $\mathbf{L2}_{\mathrm{latent}}\downarrow$
        & \textbf{IDM}$\downarrow$ \\
        \midrule
        baseline(Cosmos3)
                & 20.233 & 0.766 & 0.178
        & 42.987 & 36.780 & 0.365 & 1.108 \\
        \midrule
        embodied data only
        & 21.448 & 0.833 & 0.120 & 23.650 & 20.574 & 0.332 & 0.424 \\

        + human
        & 24.997 & 0.876 & 0.093 & \underline{15.229} & 16.980 & 0.303 & 0.391 \\

        $\Delta$ gain $\uparrow$
        & 3.549 & 0.043 & 0.027 & 8.421 & 3.594 & 0.029 & 0.033 \\

        \midrule

        embodied data + pointflow
        & 21.573 & 0.835 & 0.117 & 22.993 & 20.011 & 0.329 & 0.406 \\

        + human + pointflow
        & \underline{25.294}
        & \underline{0.880}
        & \underline{0.090}
        & 15.327
        & \underline{16.972}
        & \underline{0.227}
        & \underline{0.376} \\

        + human + stabilized pointflow
        & \textbf{25.828}
        & \textbf{0.887}
        & \textbf{0.086}
        & \textbf{14.993}
        & \textbf{15.991}
        & \textbf{0.198}
        & \textbf{0.355} \\

        $\Delta$ gain $\uparrow$
        & 4.255 & 0.052 & 0.031 & 8.000 & 4.020 & 0.131 & 0.051 \\

        \bottomrule
    \end{tabular}
    }
     \vspace{-0.25cm}
\end{table*}
\subsection{Ablation Studies}
\textbf{Effectiveness of human data.}
In the first stage of fine-tuning, we evaluate the contributions of human interaction data and our dual-branch architecture with auxiliary pointflow supervision. Compared with using embodied data alone, incorporating human data yields substantial performance gains. Further aligning human and robot motions through a unified pointflow representation leads to additional improvements. For egocentric human videos, we additionally stabilize the pointflow to suppress camera-induced motion. As shown in Fig.~\ref{fig:ab} and Table~\ref{tab:human_ab}, combining human data with stabilized pointflow yields larger $\Delta$ gains across most metrics than adding human data alone, measured against their respective embodied-data baselines.
These results suggest that unifying human and robot motion representations while suppressing camera-induced motion facilitates cross-embodiment knowledge transfer, enabling embodied world models to learn more effectively from human manipulation data.

\begin{table*}[t]
    \centering
    \caption{Quantitative ablation of GRPO for learning from failure cases.}
    \label{tab:grpo_ab}
    \resizebox{\textwidth}{!}{
    \begin{tabular}{lcccccccc}
        \toprule
        \textbf{Variant}
        & \textbf{PSNR}$\uparrow$
        & \textbf{SSIM}$\uparrow$
        & \textbf{LPIPS}$\downarrow$
        & \textbf{FVD}$\downarrow$
        & \textbf{FID}$\downarrow$
        & $\mathbf{L2}_{\mathrm{latent}}\downarrow$
        & \textbf{IDM}$\downarrow$ \\
        \midrule
        w/o GRPO
        & \underline{25.828}
        & 0.887
        & 0.086
        & {14.993}
        & {15.991}
        & 0.198
        & {0.355} \\

       failure-aware GRPO + action alignment
        & 25.826 & \underline{0.887} & \underline{0.086}
        & \underline{14.957} & 15.957 & \underline{0.198} & \underline{0.348}  \\
        failure-aware GRPO + physical plausibility
        & 25.574 & 0.862 & 0.098 & 15.307
        & \underline{15.454} & 0.209 & 0.381  \\
        failure-aware GRPO + full rewards
        & \textbf{26.077} & \textbf{0.889} & \textbf{0.082} & \textbf{14.449}
        & \textbf{14.291} &\textbf{0.193} & \textbf{0.332}  \\
        \bottomrule

    \end{tabular}
    }
            \vspace{-0.4cm}
\end{table*}

\textbf{Effectiveness of failure-aware GRPO.}
We evaluate the contribution of GRPO and its reward components in the second stage. Fig.~\ref{fig:ab} illustrates improved failure modeling, while Table~\ref{tab:grpo_ab} quantifies the complementary effects of action alignment and physical plausibility.
As shown in Fig.~\ref{fig:reward_ab} and Fig.~\ref{fig:case}, using only the physical plausibility reward tends to produce nearly static scenes or physically plausible motion that contradicts the conditioning actions. Conversely, using only the action alignment reward preserves action-consistent robot motion but compromises temporal consistency in background regions and the physical plausibility of interaction dynamics. Combining the two rewards improves action adherence while maintaining physically coherent scene dynamics. Table~\ref{tab:grpo_ab_success} further shows that GRPO improves performance on both successful and failed executions. Additional ablation studies are provided in Appendix~\ref{more_ab}.

\subsection{Policy Evaluation}
\label{app:policy_evaluation}

\begin{wrapfigure}[10]{r}{0.44\linewidth}
    \vspace{-3\baselineskip}
    \centering
    \includegraphics[width=\linewidth]{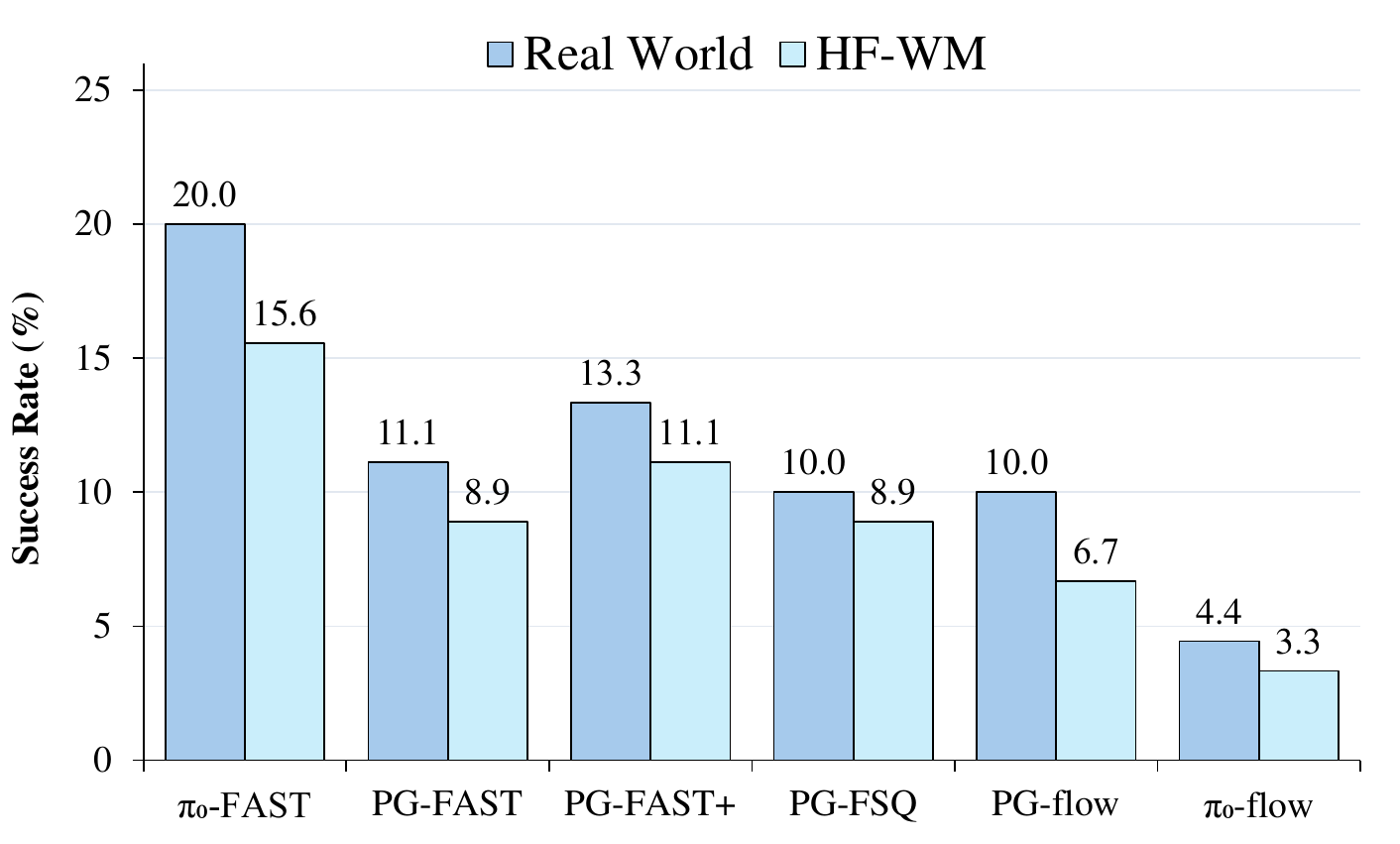}
    \caption{Real-world and HF-WM estimated success rates across six generalist robot policies.}
    \label{fig:roboarena}
\end{wrapfigure}

To assess HF-WM as a real-world policy evaluator, we select six generalist policies from the RoboArena leaderboard \citep{roboarena} and evaluate each policy on the same 90 matched execution sessions, yielding $540$ policy-session evaluations in total. For each session, HF-WM receives only the initial observation and the corresponding policy action sequence and autoregressively predicts the subsequent interaction. GPT-5 \citep{gpt5} evaluates task success from the generated videos, yielding a predicted success rate for each policy.
\mbox{The resulting} policy estimates are further compared with real-world deployment performance.
As shown in Fig.~\ref{fig:roboarena},
the predicted success rates exhibit a strong linear correlation with their real-world counterparts, with Pearson \(r=0.983\).
MMRV \citep{evaluating} measures the preservation of relative policy performance, with \(\mathrm{MMRV}=0.002\). These results show that HF-WM accurately reflects both policy performance and relative rankings, demonstrating its potential as a robot policy evaluator. Additional metrics and statistical analyses are provided in the Appendix \ref{app:policy_evaluation_ap}.

\section{Conclusion}
We presented HF-WM, an action-conditioned world model that integrates egocentric human interaction learning with failure-aware post-training. Its dual-branch architecture employs stabilized pointflow supervision to establish a shared motion representation across human and robot interactions while reducing the effects of egocentric camera motion. GRPO post-training on perturbed actions improves failure dynamics modeling through complementary action alignment and physical plausibility rewards, without paired ground-truth videos. Experiments demonstrate state-of-the-art performance in action-conditioned video generation and accurate estimation of real-world policy performance. These results highlight the value of learning from human interactions and modeling failure dynamics in extending world models beyond successful robot demonstrations.

\bibliography{paper}
\bibliographystyle{corlabbrvnat}

\clearpage
\appendix
\section{Implementation Details}
\subsection{Learning Across Embodiments: From Human to Robot}
\label{app:implementation}

\noindent\textbf{Attention mask.}
As illustrated in Fig.~\ref{fig:mask}, we retain the dual-stream attention mechanism of Cosmos 3 \citep{cosmos3}. The backbone consists of an autoregressive (AR) stream and a diffusion-model (DM) stream. AR queries attend causally within the AR stream, whereas DM queries attend bidirectionally to the AR stream and to all tokens in the DM stream, including video, skeleton, and action tokens.
To explicitly constrain the motion of the generated video, we introduce pointflow (PF) queries as an auxiliary visual-motion readout branch outside the backbone token sequence. PF queries self-attend only within the PF branch and cross-attend exclusively to intermediate video features from the DM stream. Although video, skeleton, and action tokens are all part of the DM stream, skeleton and action keys and values are explicitly masked from PF cross-attention; AR, language, and other non-video features are also inaccessible. This design prevents the PF branch from taking a shortcut through the conditioning signals and forces the pointflow objective to be grounded in the generated video representations, improving motion consistency and geometric fidelity.
The allowed attention pattern is:

\begin{figure}[h]
\centering
\makebox[\linewidth][c]{%
\begin{minipage}[c]{0.36\linewidth}
\vspace{0pt}
\centering
\includegraphics[width=\linewidth]{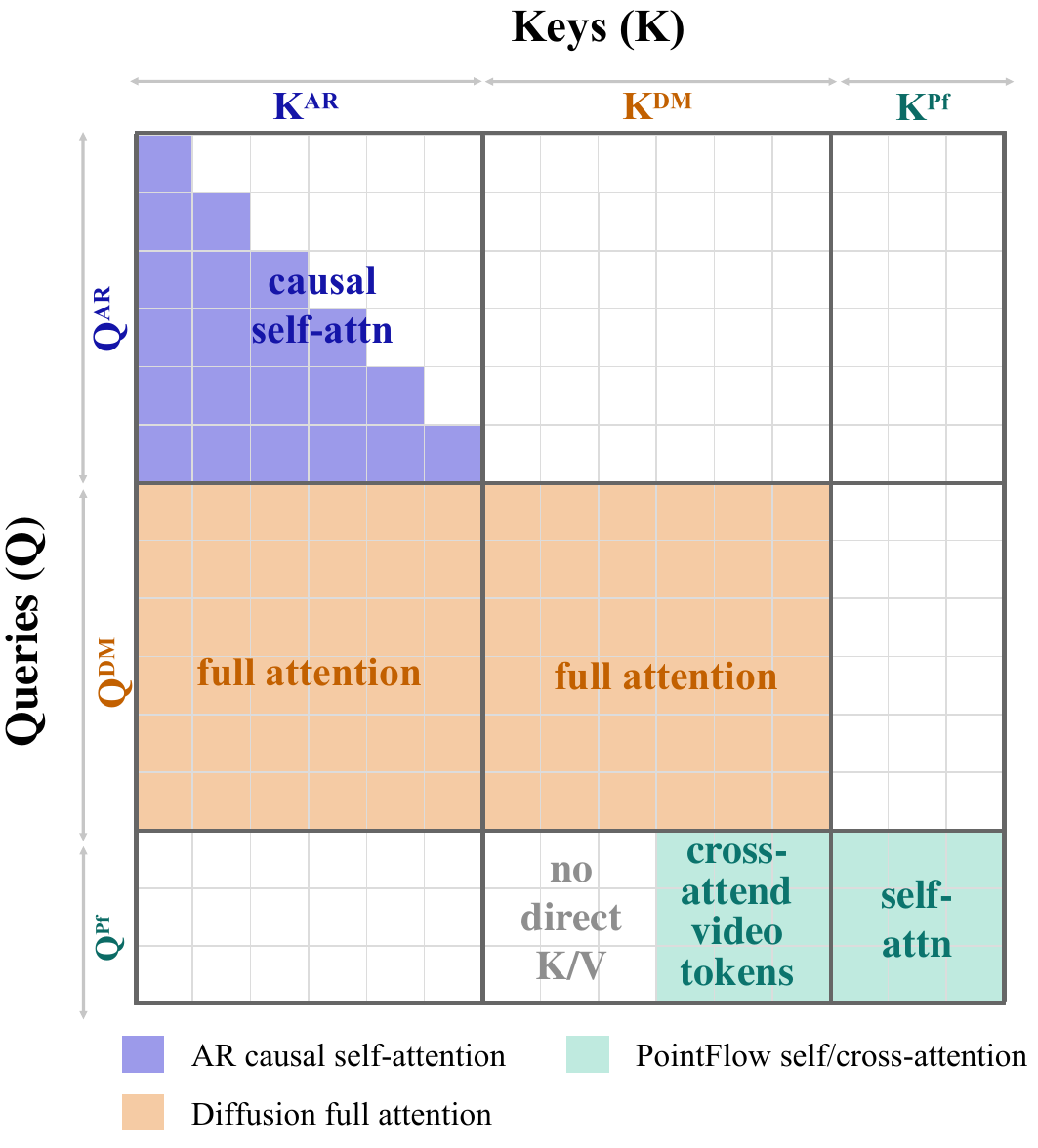}
\end{minipage}
\hspace{0.02\linewidth}%
\begin{minipage}[c]{0.61\linewidth}
\vspace{0pt}
\centering
\scriptsize
\renewcommand{\arraystretch}{1.22}
\setlength{\tabcolsep}{0pt}
\begin{tabular}{@{}>{\raggedright\arraybackslash}p{0.23\linewidth}
                @{\hspace{0.025\linewidth}}
                >{\raggedright\arraybackslash}p{0.41\linewidth}
                @{\hspace{0.025\linewidth}}
                >{\raggedright\arraybackslash}p{0.31\linewidth}@{}}
\toprule
\textbf{Query group} &
\textbf{Accessible keys and values} &
\textbf{Masking constraint} \\
\midrule
AR queries &
AR keys and values &
Causal attention\newline within AR stream \\
\midrule
DM queries &
AR keys/values and\newline video, skeleton, action &
Bidirectional attention\newline over AR and DM streams \\
\midrule
PF cross-attention &
Intermediate DM video\newline features only &
Skeleton/action/\newline language; AR and non-video\newline keys/values masked \\
\midrule
PF self-attention &
PF keys and values &
Self-attention restricted\newline to the PF branch \\
\bottomrule
\end{tabular}
\end{minipage}
}%
 \caption{HF-WM architecture and attention visibility. Left: attention mask. Right: accessible keys/values and masking constraints for AR, DM, and PF queries.}
\label{fig:mask}
\end{figure}

\textbf{Stabilized PointFlow Processing.} The main text describes our camera-motion compensation for pointflow; here, we provide additional preprocessing details. For each 81-frame video, we track \(1008=24\times42\) points on a regular grid and establish their 3D reference positions from the first frame using depths predicted by Depth Anything 3 \citep{lin2025depth}. Depth values are bilinearly sampled at the initial point locations, and the points are reprojected to subsequent frames using the ground-truth camera poses to estimate camera-induced motion. This motion is then subtracted from the original CoWTracker trajectories to obtain stabilized point trajectories.
To align pointflow supervision with the temporal resolution of the video latents, we construct relative pointflow with respect to the first frame and temporally subsample the 81-frame trajectories with a stride of 4, yielding 20 supervision time steps.

\begin{figure}[t]
    \centering
    \includegraphics[width=1.0\linewidth]{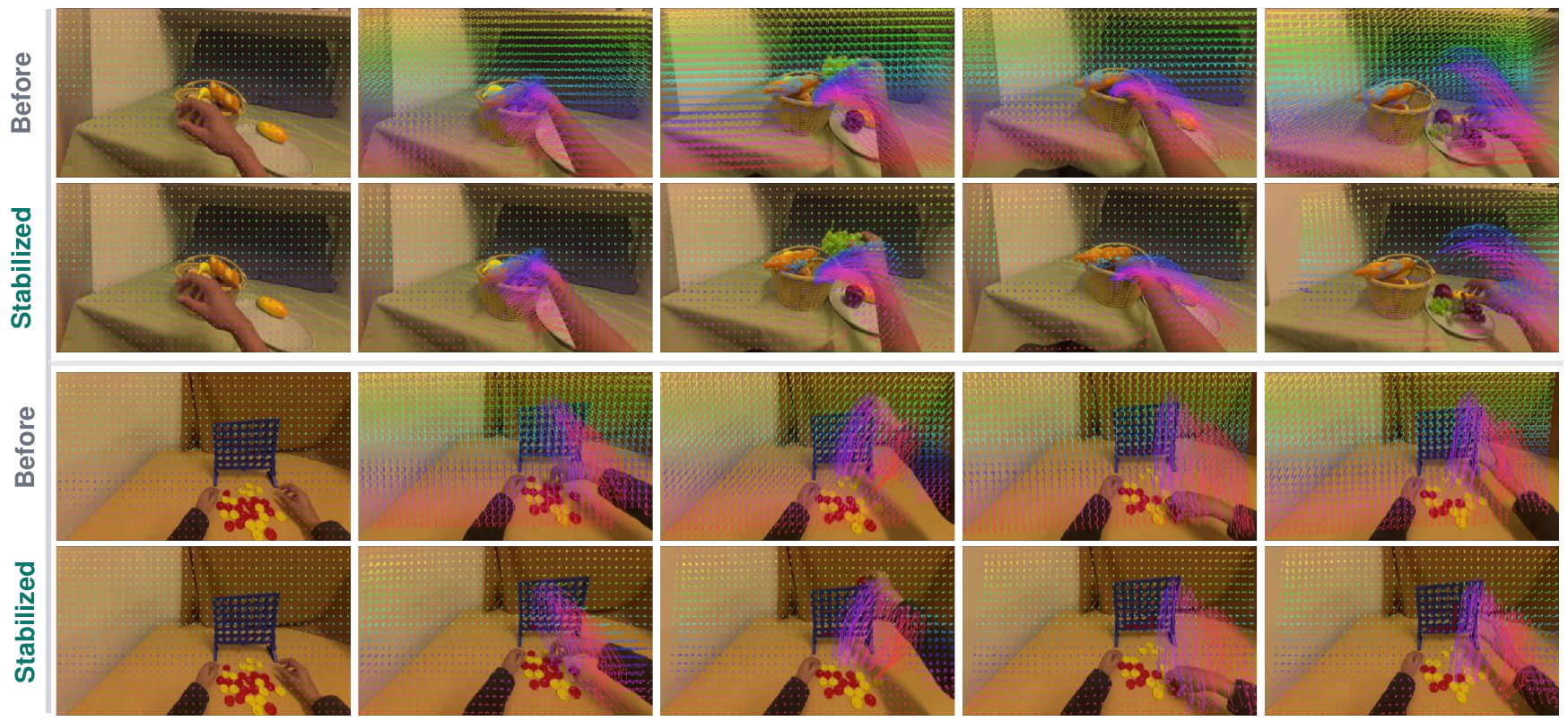}
    \caption{Visualization of pointflow stabilization on egocentric interactions. Camera-motion compensation suppresses coherent background drift while preserving hand-object interaction motion.}
    \label{fig:stab}
    \vspace{-0.2cm}
\end{figure}

As shown in Fig.~\ref{fig:stab}, static background points in the raw pointflow exhibit coherent drift under egocentric camera motion. After camera-motion compensation, this global background motion is substantially reduced, while the relative motion of the hand and manipulated objects is preserved. This indicates that the stabilization primarily removes viewpoint-induced motion rather than indiscriminately smoothing the point trajectories.

\subsection{Learning Across Outcomes: From Success to Failure}
We construct fixed-length GRPO training clips anchored to gripper-state transitions. For each episode, we identify the frame with the largest absolute change in gripper state relative to the preceding frame. Each clip contains 41 frames sampled with a stride of 2, spanning 81 source frames. We set the window start 12 source frames before the detected transition and clip it to the valid range to keep the full window within the episode. This event-anchored sampling captures pre-transition context and subsequent interaction dynamics without requiring the transition to lie at the window midpoint.
We then classify each sampled window by its gripper-state transition pattern to guide subsequent action perturbations. We apply a length-3 median filter to suppress transient fluctuations and identify stable states using dual-threshold hysteresis: values at or below 0.25 indicate a closed state, values at or above 0.75 indicate an open state, and intermediate values do not trigger state transitions. Starting from the first stable state, we track the initial transition direction and any subsequent reversal to distinguish four patterns: closure without reopening (\texttt{close\_hold}), closure followed by reopening (\texttt{close\_reopen}), opening without reclosing (\texttt{open\_hold}), and opening followed by reclosing (\texttt{open\_close}). Windows without valid transitions are classified by their overall gripper-value range as continuously open (\texttt{remain\_open}), continuously closed (\texttt{remain\_closed}), or indeterminate (\texttt{partial\_hold}). This seven-way classification determines eligibility for gripper perturbations and guides the choice between gripper and geometric perturbations.
After gripper-event classification, we sample training windows according to predefined class quotas, allocating 75\% to the four classes with explicit gripper-state transitions and 25\% to the remaining three classes. For each window, we select the unperturbed route (\texttt{none}) with probability 50\%, retaining the original actions and corresponding video; otherwise, we enter the counterfactual perturbation branch. This balanced sampling aims to prevent overemphasis on perturbed executions while preserving the modeling of original executions as the model learns failure-related dynamics.

The counterfactual branch selects perturbations according to the event class. For windows with explicit state transitions, gripper and geometric perturbations are sampled with probabilities of 70\% and 30\%, respectively. This yields within-class routing probabilities of 50\% for no perturbation, 35\% for gripper perturbations, and 15\% for geometric perturbations. The specific gripper operation depends on the event pattern, as summarized in Table~\ref{tab:perturbation_routing}. Within the \texttt{open\_hold} gripper branch, reclosing (\texttt{reclose}) and suppression of gripper-state changes (\texttt{suppress}) are selected with probabilities of 25\% and 75\%, respectively. For windows without valid state transitions, the counterfactual branch applies only geometric perturbations.
Geometric perturbations are sampled uniformly across one rotation route and four translation routes. Each geometric route therefore has a within-class selection probability of 3\% for classes with explicit state transitions and 10\% for those without valid transitions.

During route execution, gripper perturbations modify only the gripper-state sequence, preserving the arm joint trajectory and end-effector poses without requiring inverse kinematics (IK). Geometric perturbations must satisfy both visual and kinematic constraints.
For each geometric candidate, we compute end-effector poses from the original joint states and construct a target trajectory using local $\mathrm{SE}(3)$ perturbations. A cosine ramp smoothly increases the translational and rotational offsets from zero to their target magnitudes, mitigating abrupt trajectory changes. Before solving IK, we verify sufficient gripper visibility in both the source skeleton and the unperturbed baseline.
We solve IK via bounded nonlinear least squares under the URDF-specified joint limits, minimizing normalized end-effector position and orientation errors. Each subsequent solve is warm-started from the preceding solution. A candidate passes kinematic validation only if all joint values are finite and within limits, position and orientation residuals fall below their respective thresholds, and joint changes between consecutive states satisfy continuity constraints.
For kinematically feasible candidates, we apply forward kinematics to the resulting joint trajectory and re-render the skeleton and gripper using the camera intrinsics and extrinsics. We verify that the end effector remains within the valid image region and the gripper remains visible. Only candidates satisfying both visual and kinematic constraints are accepted. If IK or action reconstruction fails, or re-rendering reveals insufficient gripper visibility or anomalous displacement, we restore the original inputs and try another eligible route.

\begin{table*}[t]
    \centering
    \caption{
        Gripper event patterns, target sampling shares, and perturbation
        routing probabilities. Target shares are relative to all training
        samples; routing probabilities are conditional on the event class
        and sum to 100\% per row.
    }
    \label{tab:perturbation_routing}

    \normalfont\small
    \setlength{\tabcolsep}{3pt}
    \setlength{\gripTableWidth}{
        \dimexpr\textwidth-10\tabcolsep\relax
    }
    \renewcommand{\arraystretch}{1}

    \begin{tabular}{@{}
        G{0.16\gripTableWidth}
        G{0.30\gripTableWidth}
        H{0.09\gripTableWidth}
        H{0.075\gripTableWidth}
        G{0.245\gripTableWidth}
        H{0.13\gripTableWidth}
        @{}}
        \toprule

        \textbf{Event class}
        & \textbf{Gripper event pattern}
        & \textbf{Target}\newline\textbf{share}
        & \textbf{None}
        & \textbf{Gripper}\newline\textbf{perturbation}
        & \textbf{Geometric}\newline\textbf{total} \\

        \midrule

        close\_hold
        & \gripopen{} $\rightarrow$ \gripclosed{} (hold)
        & 18.75\%
        & 50\%
        & reopen: 35\%
        & 15\% \\

        close\_reopen
        & \gripopen{} $\rightarrow$ \gripclosed{}
          $\rightarrow$ \gripopen{}
        & 18.75\%
        & 50\%
        & force\_closed: 35\%
        & 15\% \\

        open\_hold
        & \gripclosed{} $\rightarrow$ \gripopen{} (hold)
        & 18.75\%
        & 50\%
        & reclose: 8.75\%\newline
          suppress: 26.25\%
        & 15\% \\

        open\_close
        & \gripclosed{} $\rightarrow$ \gripopen{}
          $\rightarrow$ \gripclosed{}
        & 18.75\%
        & 50\%
        & force\_open: 35\%
        & 15\% \\

        \specialrule{0.35pt}{0pt}{0pt}

        remain\_open
        & \gripopen{} (hold)
        & 12\%
        & 50\%
        & Not applied
        & 50\% \\

        partial\_hold
        & \gripuncertain{}
        & 8\%
        & 50\%
        & Not applied
        & 50\% \\

        remain\_closed
        & \gripclosed{} (hold)
        & 5\%
        & 50\%
        & Not applied
        & 50\% \\

        \bottomrule
    \end{tabular}

    \par\vspace{3pt}
      \vspace{-0.4cm}
\end{table*}

\label{more_grpo}
\subsubsection{Action Alignment Reward}
\begin{figure}[!t]
    \centering
    \includegraphics[width=1.0\linewidth]{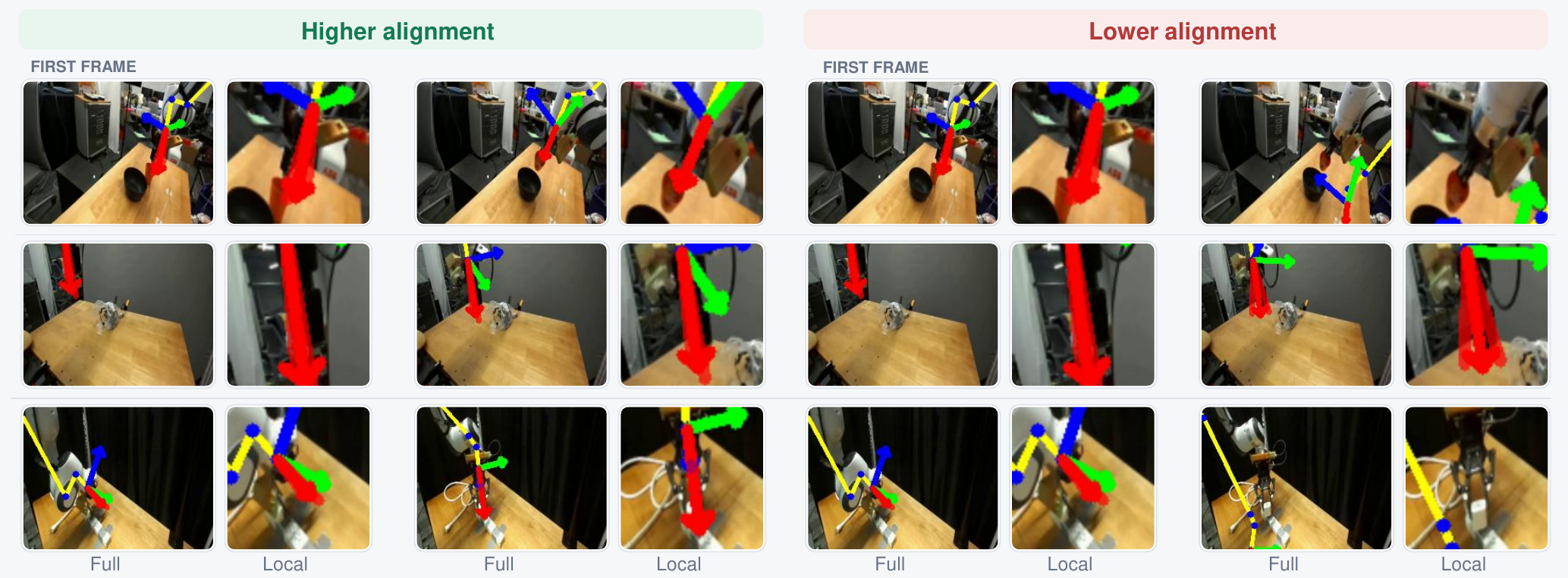}
    \caption{\textbf{Qualitative examples for the action alignment reward.} Each row compares video-action pairs with higher (left) and lower (right) alignment. Action-skeleton overlays in full-frame views and end-effector crops highlight differences in consistency between the specified actions and observed robot motion.}
    \label{fig:action_alignment_reward}
      \vspace{-0.5cm}
\end{figure}

As illustrated in Fig.~\ref{fig:action_alignment_reward}, our action alignment reward model uses a global-local dual-branch architecture to assess consistency between generated videos and conditioning actions at complementary visual scales. Both branches process 41-frame sequences aligned with GRPO rollouts. The global branch jointly encodes RGB frames and their action-skeleton overlays at $224\times224$ resolution to capture overall robot motion. The local branch extracts $80\times80$ end-effector crops from the overlays and resizes them to $160\times160$ to capture fine-grained motion. Independent convolutional encoders extract per-frame visual features from each branch, which are fused and projected into unified frame-level representations. These representations are augmented with temporal positional encodings and processed by a three-layer temporal Transformer to jointly model global motion trajectories and local end-effector dynamics.
We construct training samples spanning different alignment levels through controlled action perturbations. For each real interaction clip, we keep the original RGB video fixed, perturb the action sequence with varying types and magnitudes, and re-render the corresponding skeletons using the robot's URDF and forward kinematics. Each action candidate is paired with the same video, focusing supervision on alignment differences between conditioning actions and observed motion.

Perturbations cover arm pose, wrist rotation, temporal shifts, and gripper states. Gripper perturbations include forced opening or closure, state inversion, delayed transitions, and reopening after closure. Pose and temporal perturbations span multiple magnitudes, and combined perturbations extend coverage from minor deviations to severe misalignment. Since few-step denoising in GRPO may degrade local image quality, we additionally apply visual corruptions, such as blur and noise, to end-effector regions while preserving the actions and skeleton trajectories. These augmentations aim to reduce the influence of appearance quality on alignment assessment.
Training samples are organized into preference pairs according to relative alignment, including comparisons between original and perturbed actions as well as between perturbed actions with different degrees of deviation. The preferred candidate is more consistent with the observed video but need not be fully correct. Thus, supervision distinguishes not only correct from incorrect action conditions but also mild from severe misalignment.

Each preference pair is supervised with a continuous target score gap, determined by predefined rules based on perturbation type and magnitude. Gripper deviations are quantified by differences in openness from the original trajectory over non-conditioning frames. For pose, rotation, and temporal perturbations, target gaps are determined by joint offsets, rotation angles, and temporal shifts, respectively. Identical or visually indistinguishable inputs receive a zero target gap to avoid amplifying unobservable action differences into reward differences. This graded supervision captures both relative ordering and the extent of alignment differences, rather than binary correctness alone.

\begin{figure}[!tp]
    \centering
    \includegraphics[width=0.96\linewidth]{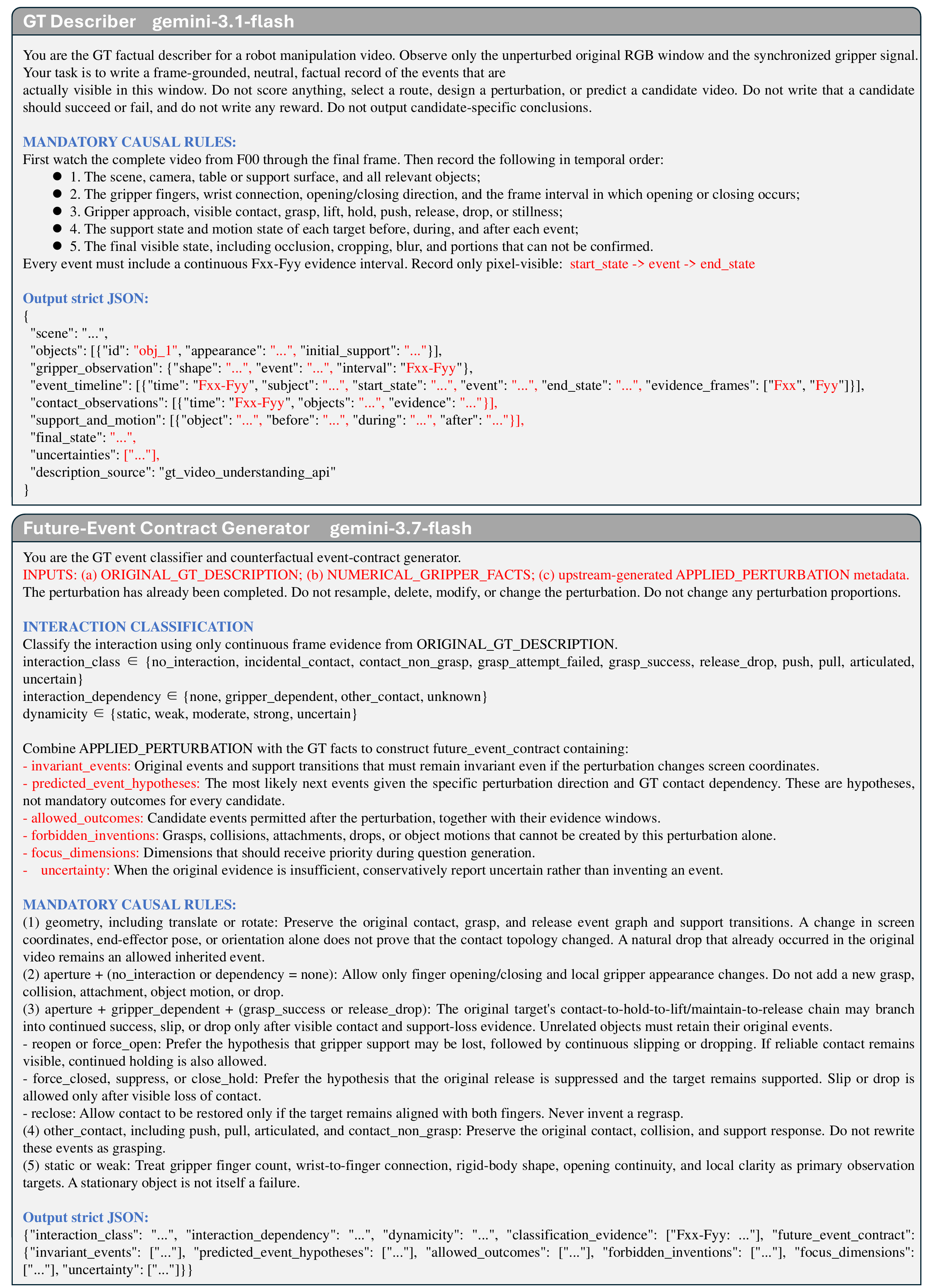}
    \caption{Prompt template for GT Describer and Future-Event Contract Generator.}
    \label{fig:prompt}
\end{figure}

\subsubsection{Physical Plausibility Reward}
\label{ppr}
We extract interaction events, contact relations and support conditions from the original RGB video, annotating the corresponding evidence frame intervals. This establishes an observational baseline independent of perturbation routing and candidate outcomes for subsequent counterfactual assessment. Combining these observations with perturbation metadata from the preceding classification and routing stage, we construct counterfactual event descriptions based on interaction type, gripper dependence, and scene dynamics. These descriptions specify event relations to preserve, permissible state changes, and uncertainty where evidence is insufficient, defining a set of physically admissible future outcomes.

Specifically, translational or rotational perturbations should generally preserve the original event graph of contact, grasping, release, and support. In contact-free scenes, perturbations to gripper opening and closing should affect only the gripper itself. For gripper-dependent grasps, state changes such as slippage or falling require visible evidence of contact or support loss in the candidate video, rather than being inferred solely from action intent or gripper signals.

Building on these observations and counterfactual event descriptions, we assess physical plausibility through tiered binary question answering \citep{abot}, using Gemini 3.7 Flash for both question generation and candidate evaluation. Without access to candidate videos, the question generator constructs and freezes a checklist shared by all candidates in a group. Tier-1 comprises three critical checks targeting severe physical or structural violations, such as object interpenetration, unsupported floating, and major inconsistencies in robot geometry. Tier-2 comprises five fine-grained checks spanning four dimensions: ``physical consistency,'' assessing object persistence, motion continuity, and plausible dynamics under the observed support conditions; ``robot structure,'' examining the geometric integrity and temporal consistency of the arm and gripper; ``contact causality,'' assessing whether object motion is consistent with visible contact, grasping, release, and support transitions; and ``visual quality,'' checking for artifacts that obscure object states or interaction events. These checks are tailored to scene-specific objects and counterfactual interaction conditions, rather than requiring the original demonstration outcome to be reproduced. Task failure alone is therefore not treated as a physical violation.
In a separate model call, the evaluator answers each question with \texttt{yes} (pass) or \texttt{no} (fail) based on evidence from consecutive frames of the candidate video. The physical plausibility score is the fraction of the eight checks passed. Any Tier-1 failure triggers hard rejection, regardless of the Tier-2 score, preventing favorable assessments on other dimensions from compensating for a severe physical or structural violation.
\vspace{-0.2cm}
\section{More Experimental Details}
\label{more_ex}
\subsection{Training Details}
In the first stage of supervised fine-tuning, we adopt and empirically validate the human data injection strategy from OSCAR \citep{oscar}. We first warm up the model on robot only embodied data and then jointly train with egocentric human interaction data, which yields larger gains than mixing the two data sources from the start.
We additionally find that text conditioning degrades action following and physical consistency in action-conditioned world models, as generated dynamics may follow textual semantics rather than the causal effects of the specified actions. To mitigate this interference, we use fixed text prompts throughout training and evaluation: ``A person performs a hand manipulation task.'' for human interaction samples and ``A robot performs a manipulation task.'' for robot interaction samples. This keeps the textual condition consistent and encourages the model to focus on action-driven dynamics.

During GRPO training, we follow the efficiency-oriented setting adopted in FlowFactory \citep{flowfactory}, where effective optimization can be achieved with shorter clips, lower spatial resolution, and fewer denoising steps. Accordingly, for each sample, we generate four candidate rollouts at a resolution of \(256\times256\) using 12 denoising steps, with SDE noise of strength \(0.8\) injected at the first step. The first observation frame is provided as the visual condition, from which each candidate predicts 40 future frames. To preserve sufficient interaction dynamics within this compact temporal window, we select an 81-frame action segment around each gripper state transition and uniformly subsample it to 40 action steps aligned with the predicted frames.
This design retains the key motion and contact changes around gripper interactions while keeping GRPO training computationally efficient. We combine the action alignment and physical plausibility rewards with weights of \(0.4\) and \(0.6\), respectively.
\subsection{Comparison with Baselines}
As shown in Table~\ref{tab:comparison_success}, existing methods exhibit a clear performance drop on failed executions, with particularly pronounced degradation in IDM. This suggests that, under failure-inducing actions, these models remain biased toward the distribution of successful trajectories and tend to generate future states that resemble successful outcomes, thereby reducing fidelity to the conditioned actions. In contrast, HF-WM maintains closely comparable performance across successful and failed executions, indicating more consistent modeling of action consequences across different execution outcomes. (We use the Wan2.2 VAE to obtain the $\mathbf{L2}_{\mathrm{latent}}$)

\begin{table*}[t]
    \centering
    \caption{
    Quantitative comparison with baselines on successful and failed executions.
    Each entry reports Success / Failure.
    \textbf{Bold} / \underline{underline} denote the best / second-best results.
    }
    \label{tab:comparison_success}
    \resizebox{\textwidth}{!}{
    \begin{tabular}{lcccccccc}
        \toprule
        \textbf{Method (Success/Failure)}
        & \textbf{PSNR}$\uparrow$
        & \textbf{SSIM}$\uparrow$
        & \textbf{LPIPS}$\downarrow$
        & \textbf{FVD}$\downarrow$
        & \textbf{FID}$\downarrow$
        & $\mathbf{L2}_{\mathrm{latent}}\downarrow$
        & \textbf{IDM}$\downarrow$ \\
        \midrule
        Cosmos-Predict2.5 \citep{cosmos}
        & 19.869/17.878 & 0.755/0.736 & 0.268/0.263
        & 42.001/78.237 & 47.525/60.691 & 0.449/0.523 & 1.191/1.344 \\

        TesserAct \citep{tesseract}
        & 18.192/16.574 & 0.762/0.751 & 0.245/0.280
        & 59.452/61.673 & 59.053/60.724 & 0.517/0.528 & 1.522/1.804 \\

        Ctrl-World \citep{ctrl}
        & 21.746/21.620 & 0.788/0.752 & 0.301/0.309
        & 32.247/41.585 & 61.721/67.608 & 0.406/0.429 & 0.887/1.011 \\

        Cosmos 3 \citep{cosmos3}
        & 21.172/19.293 & 0.815/0.716 & 0.142/0.213
        & 38.512/47.461 & 34.205/39.354 & 0.349/0.381 & 1.101/1.114 \\

        OSCAR \citep{oscar}
        & \underline{23.674/23.040}
        & \underline{0.860/0.820}
        & 0.118/\underline{0.165}
        & 29.638/33.992
        & \underline{21.297/35.185}
        & \underline{0.243/0.340}
        & \underline{0.445/0.712} \\

        Kinema4D \citep{kinema4d}
        & 20.630/19.400 & 0.834/0.773 & \underline{0.117}/0.197
        & \underline{27.155/33.866} & 23.495/38.285 & 0.270/0.381 & 0.904/0.923 \\
        \midrule
        \method{} (\textbf{Ours})
        & \textbf{26.321/25.833}
        & \textbf{0.892/0.886}
        & \textbf{0.066/0.098}
        & \textbf{13.350/15.547}
        & \textbf{14.310/14.271}
        & \textbf{0.176/0.210}
        & \textbf{0.328/0.336} \\
        \bottomrule
    \end{tabular}
    }
\end{table*}

\begin{table*}[t]
    \centering
    \caption{Ablation study on the feature extraction layers of the pointflow branch.(SFT)}
    \label{tab:layer_ab}
    \resizebox{\textwidth}{!}{
    \begin{tabular}{lccccccc}
        \toprule
        \textbf{Variant}
        & \textbf{PSNR}$\uparrow$
        & \textbf{SSIM}$\uparrow$
        & \textbf{LPIPS}$\downarrow$
        & \textbf{FVD}$\downarrow$
        & \textbf{FID}$\downarrow$
        & $\mathbf{L2}_{\mathrm{latent}}\downarrow$
        & \textbf{IDM}$\downarrow$ \\
        \midrule

        mid [12,14,16,18]
        & 25.756
        & \underline{0.886}
        & 0.087
        & 15.163
        & \underline{16.031}
        & \underline{0.201}
        & 0.377 \\

        back [20,22,24,26]
        & 25.689
        & 0.885
        & 0.087
        & 15.326
        & 16.361
        & 0.202
        & 0.393 \\

        uniform [8,14,20,26]
        & \underline{25.761}
        & \underline{0.886}
        & \underline{0.086}
        & \underline{15.047}
        & 16.211
        & \underline{0.201}
        & \underline{0.357} \\

        front [4,6,8,10]
        & \textbf{25.828}
        & \textbf{0.887}
        & \textbf{0.086}
        & \textbf{14.993}
        & \textbf{15.991}
        & \textbf{0.198}
        & \textbf{0.355} \\

        \bottomrule
    \end{tabular}

    }
\end{table*}
\subsection{Ablation Studies}
\label{more_ab}
\subsubsection{Ablation on PointFlow Feature Layers}
To study the effect of where the auxiliary pointflow branch is attached, we compare intermediate video features extracted from different backbone depths as inputs to pointflow supervision. Specifically, we evaluate four configurations: front layers [4, 6, 8, 10], middle layers [12, 14, 16, 18], back layers [20, 22, 24, 26], and uniformly distributed layers [8, 14, 20, 26]. As shown in Table~\ref{tab:layer_ab}, the front-layer configuration performs best, followed by uniform sampling, while the middle- and back-layer variants show progressively lower performance. This trend suggests that pointflow supervision benefits more from shallower representations that retain fine-grained spatial structure and motion cues. Uniform sampling still captures complementary information across feature hierarchies, but deeper representations provide less direct motion information. We therefore use [4, 6, 8, 10] for the pointflow branch in our final model.

\subsubsection{Ablation on GRPO Rewards}
\begin{figure}[!t]
    \centering
    \includegraphics[width=1.0\linewidth]{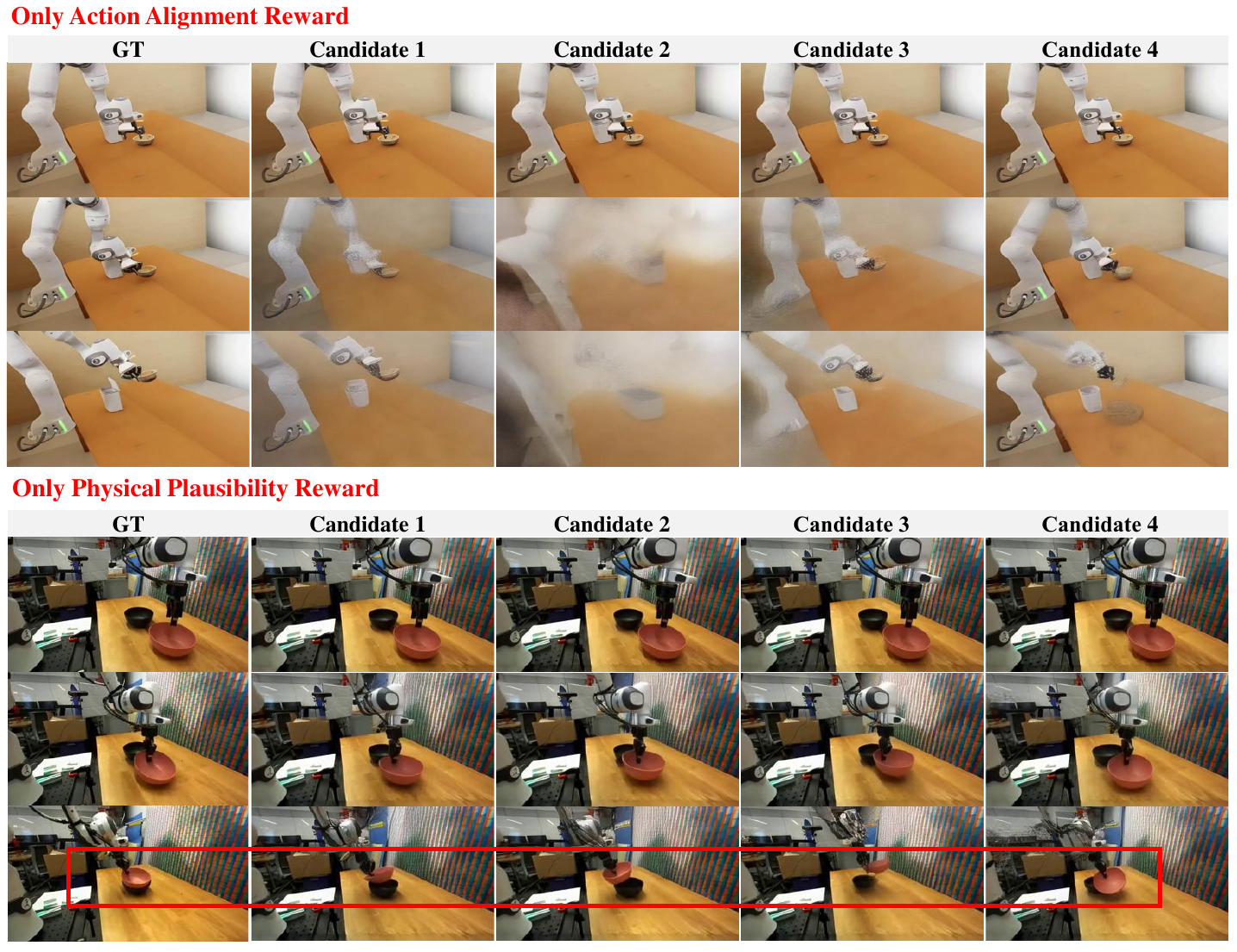}
    \caption{\textbf{Intermediate rollout examples during GRPO training with individual reward components.} Each panel shows corresponding frames from a ground-truth video and four candidate rollouts. Top: With only the action alignment reward, some candidates exhibit physically implausible interaction dynamics and degraded background quality. Bottom: With only the physical plausibility reward, the generated motion appears physically plausible but deviates from the conditioning actions. (\textbf{Note.} Variation across candidates arises from stochastic exploration during GRPO training. Their relatively lower visual fidelity is a consequence of the reduced denoising steps adopted for training efficiency.)}
    \label{fig:reward_ab}
    \vspace{-0.2cm}
\end{figure}

\begin{figure}[!t]
    \centering
    \includegraphics[width=1.0\linewidth]{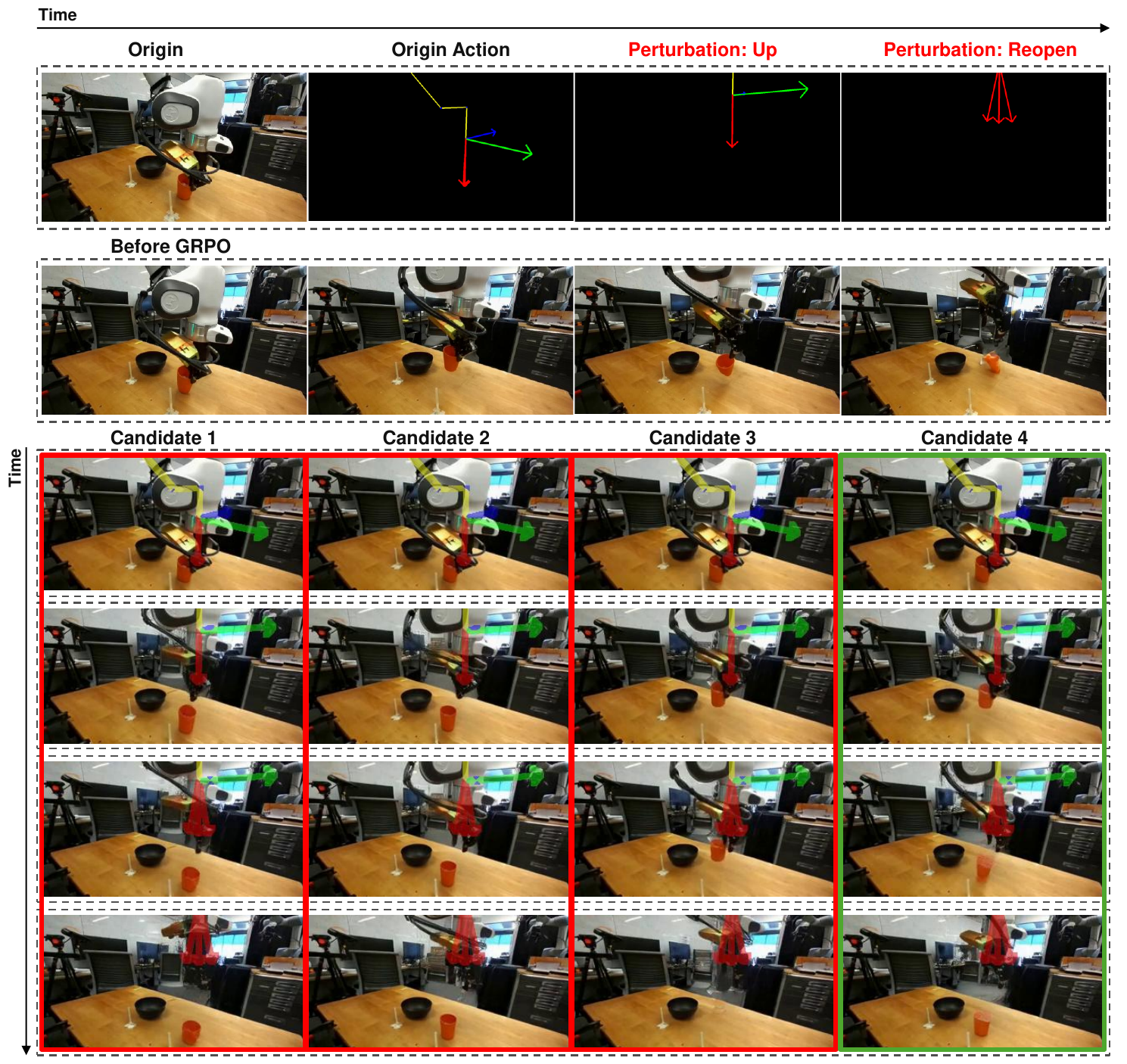}
    \caption{\textbf{Qualitative case of GRPO post-training under action perturbations.} Starting from the original action skeleton, we introduce two counterfactual perturbations, \textit{Up} and \textit{Reopen}. Before GRPO, the model fails to faithfully reflect the perturbed actions and produces physically implausible outcomes, such as the cup remaining unnaturally suspended instead of falling after release. After GRPO, the generated candidates better follow the perturbed actions and exhibit more physically plausible object dynamics.}
    \label{fig:case}
    \vspace{-0.4cm}
\end{figure}

Table~\ref{tab:grpo_ab_success} complements the aggregate evaluation in the main text by reporting each reward configuration separately on successful and failed executions. Either reward alone improves some metrics but does not consistently improve both generation quality and action following. In contrast, combining the two rewards improves performance on both subsets(Fig.~\ref{fig:reward_ab}).
As shown in Fig.~\ref{fig:case}, GRPO yields more physically plausible predictions of object dynamics and failure outcomes under perturbed actions. These results suggest that reward-guided post-training helps address the limited coverage of failure dynamics in successful demonstrations.
\begin{minipage}{\linewidth}
    \centering
    \captionof{table}{Ablation study of GRPO on successful and failed outcomes.}
    \label{tab:grpo_ab_success}
    \resizebox{\textwidth}{!}{
    \begin{tabular}{lcccccccc}
        \toprule
        \textbf{Variant (Success/Failure) }
        & \textbf{PSNR}$\uparrow$
        & \textbf{SSIM}$\uparrow$
        & \textbf{LPIPS}$\downarrow$
        & \textbf{FVD}$\downarrow$
        & \textbf{FID}$\downarrow$
        & $\mathbf{L2}_{\mathrm{latent}}\downarrow$
        & \textbf{IDM}$\downarrow$ \\
        \midrule
        w/o GRPO
        & 26.165/\underline{25.490} & 0.891/\underline{0.883} & 0.068/\underline{0.104} & 13.691/16.295
        & 14.989/16.993 & 0.182/0.214 & 0.334/0.376  \\
        GRPO+action alignment
        & 26.169/25.482 & 0.891/0.882 & 0.068/\underline{0.104} & 13.549/16.365
        & 14.983/16.930 & 0.182/\underline{0.213} & 0.330/\underline{0.365}  \\
        GRPO+physical plausibility
        & 25.870/25.278 & 0.887/0.837 & 0.071/0.125 & 14.674/\underline{15.940}
        & 14.951/15.957 & 0.190/0.228 & 0.363/0.398 \\
        GRPO w/o action perturbations
        & \underline{26.319}/25.434 & \textbf{0.893}/0.878 & \textbf{0.066}/0.106 & \underline{13.381}/16.143
        & \underline{14.321}/\underline{15.736} & \underline{0.179}/0.221 & \textbf{0.328}/0.374  \\
        GRPO (full)
        & \textbf{26.321}/\textbf{25.833} & \underline{0.892}/\textbf{0.886} & \textbf{0.066}/\textbf{0.098} & \textbf{13.350}/\textbf{15.547}
        & \textbf{14.310}/\textbf{14.271} & \textbf{0.176}/\textbf{0.210} & \textbf{0.328}/\textbf{0.336}  \\
        \bottomrule
    \end{tabular}
    }
\end{minipage}

\enlargethispage{\baselineskip}
\section{Downstream Task Performance Evaluation}
\subsection{Additional Analysis of Policy Evaluation}
\label{app:policy_evaluation_ap}

We complement Sec.~\ref{app:policy_evaluation} with an analysis of success-rate estimation and policy ranking consistency. Table~\ref{tab:metrics} summarizes the results for $K=6$ policies evaluated on $N=90$ matched execution sessions. For policy $i$, the real-world and predicted success rates are $s_i=n_i/N$ and $\hat{s}_i=\hat{n}_i/N$, where $n_i$ counts successful real-world executions and $\hat{n}_i$ counts successful outcomes identified by GPT-5 in the HF-WM-generated videos.

Pearson correlation measures the linear association between predicted and real-world success rates, while the mean absolute error $\mathrm{SISR}_{\Delta}$ quantifies their numerical discrepancy:
\begin{equation}
\begin{gathered}
r
=
\frac{
\sum_{i=1}^{K}(s_i-\mu_s)(\hat{s}_i-\mu_{\hat{s}})
}{
\sqrt{\sum_{i=1}^{K}(s_i-\mu_s)^2}
\sqrt{\sum_{i=1}^{K}(\hat{s}_i-\mu_{\hat{s}})^2}
}
\approx 0.9826,\\
\mathrm{SISR}_{\Delta}
=
\frac{100}{K}\sum_{i=1}^{K}|\hat{s}_i-s_i|
\approx 2.41\,\mathrm{pp},
\end{gathered}
\end{equation}
where $\mu_s=K^{-1}\sum_{i=1}^{K}s_i$ and $\mu_{\hat{s}}=K^{-1}\sum_{i=1}^{K}\hat{s}_i$ are the mean real-world and predicted success rates across policies, respectively, and $\mathrm{pp}$ denotes percentage points. These results show that the predicted success rates closely track real-world performance, with a mean absolute deviation of 2.41 percentage points.

To assess whether these estimates preserve relative policy performance, we rank policies by their real-world and predicted success rates in descending order. Let $R_i$ and $\hat{R}_i$ denote the resulting ranks, with ties assigned their average rank. Spearman's $\rho$ measures agreement between the rankings, whereas Kendall's $\tau_b$ measures pairwise ordering consistency with tie correction:
\begin{equation}
\begin{gathered}
\rho
=
\operatorname{Corr}\!\left(
(R_i)_{i=1}^{K},(\hat{R}_i)_{i=1}^{K}
\right)
\approx 0.9559,\\
\tau_b
=
\frac{C-D}{
\sqrt{(C+D+T_s)(C+D+T_{\hat{s}})}
}
\approx 0.9286.
\end{gathered}
\end{equation}
Here, $\operatorname{Corr}$ denotes Pearson correlation; $C$ and $D$ count concordant and discordant policy pairs; and $T_s$ and $T_{\hat{s}}$ count pairs tied only in the real-world or predicted success rates, respectively. Both rank correlations indicate strong agreement with the real-world policy ordering.

\begin{figure}[p]
\centering
\includegraphics[width=\textwidth]{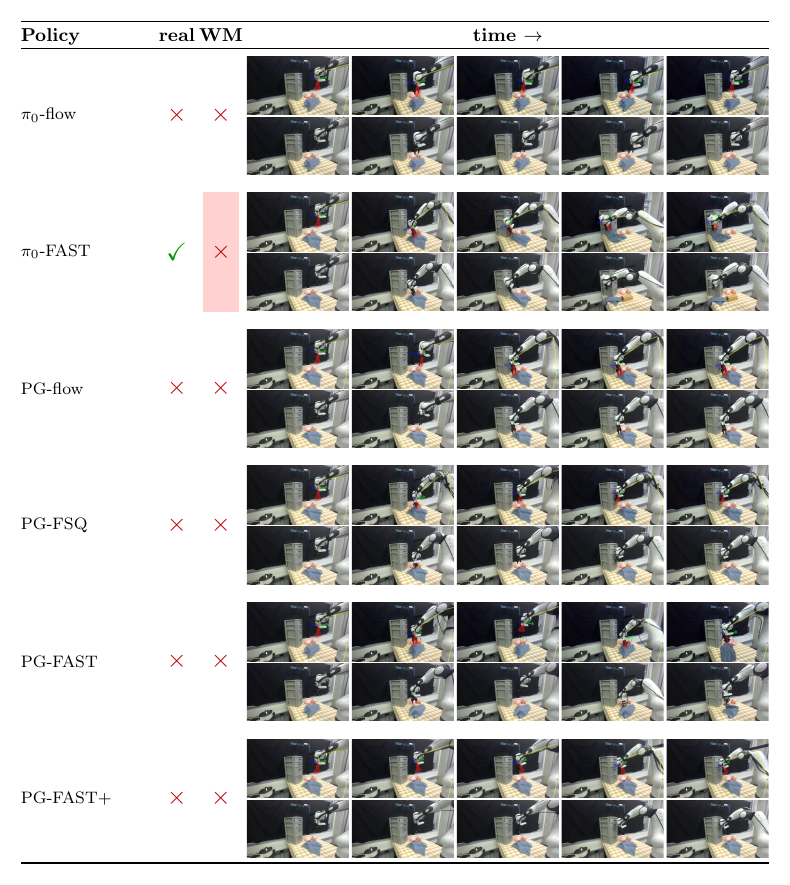}
\caption{Uncover the wooden block.
Each strip shows the world-model rollout (top) and the real-robot video (bottom) at six time steps.
The \textit{real} column is the RoboArena human success label; the WM column is the world-model verdict (Sec. \ref{app:policy_evaluation}).
A red cell marks disagreement. On this task 1 of 6 policies succeeds on the real robot.}
\label{fig:roboarena_wooden_block}
\end{figure}

\begin{figure}[p]
\centering
\includegraphics[width=\textwidth]{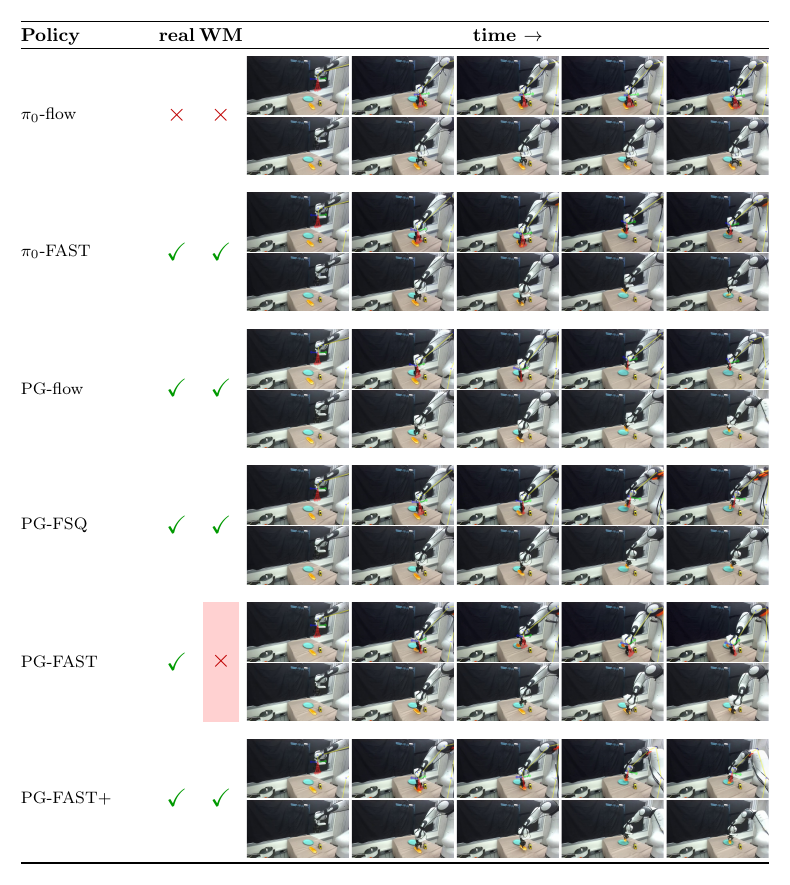}
\caption{Put the food on the plate.
Each strip shows the world-model rollout (top) and the real-robot video (bottom) at six time steps.
The \textit{real} column is the RoboArena human success label; the WM column is the world-model verdict (Sec. \ref{app:policy_evaluation}).
A red cell marks disagreement. On this task 5 of 6 policies succeed on the real robot.}
\label{fig:roboarena_food_plate}
\end{figure}

To account for the magnitude of the performance differences involved in ranking discrepancies, we further report the mean maximum rank violation (MMRV), which accounts for real-world performance gaps when measuring ranking errors. Following the official \mbox{SIMPLER implementation~\citep{evaluating}}, we compute
\begin{equation}
\begin{gathered}
v_{ij}
=
|s_i-s_j|\,
\mathbf{1}\!\left[
(\hat{s}_i>\hat{s}_j)\neq(s_i>s_j)
\right],\\
\mathrm{MMRV}
=
\frac{1}{K}\sum_{i=1}^{K}\max_{j\neq i}v_{ij}
\approx 0.002,
\end{gathered}
\end{equation}
where $\mathbf{1}[\cdot]$ denotes the indicator function and $v_{ij}$ measures a pairwise ordering violation weighted by the real-world success-rate gap. MMRV averages the largest violation for each policy. In our evaluation, the nonzero value arises from a predicted tie between PG-FAST and PG-FSQ, whose real-world success counts differ by one. Together with the high rank correlations, this result indicates that HF-WM closely preserves relative policy performance. Additional policy evaluation results are provided in Figs.~\ref{fig:roboarena_wooden_block} and~\ref{fig:roboarena_food_plate}.

\begin{table}[t]
    \centering
    \caption{HF-WM policy evaluation on 90 matched sessions.}
    \label{tab:metrics}
    \small
    \setlength{\tabcolsep}{8pt}
    \renewcommand{\arraystretch}{1.15}
    \begin{tabular}{@{}ccccc@{}}
        \toprule
        Pearson $r\uparrow$
        & Spearman $\rho\uparrow$
        & Kendall $\tau_b\uparrow$
        & $\mathrm{SISR}_{\Delta}\downarrow$ (pp)
        & MMRV $\downarrow$ \\
        \midrule
        $0.9826$
        & $0.9559$
        & $0.9286$
        & $2.41$
        & $0.002$ \\
        \bottomrule
    \end{tabular}
\end{table}

\subsection{Test-Time Scaling for Policy Inference}
We further evaluate the test-time scaling capability of HF-WM in downstream closed-loop manipulation. Experiments are conducted on all 90 tasks in LIBERO-90 using a shared $\pi_{0.5}$ policy as the base policy. To adapt HF-WM to the LIBERO domain \citep{liu2023libero}, we first fine-tune it on the LIBERO training data. Following the training setup of SC3-Eval \citep{sc3}, we use 24 frame video clips during fine-tuning to better preserve the long-horizon temporal prior of Cosmos 3 \citep{cosmos3}, while evaluating only the first 16 future steps at test time.

As shown in Fig.~\ref{fig:tts}, at each decision step, we sample four candidate action chunks from the same $\pi_{0.5}$ \citep{pi05} policy and use HF-WM to predict their corresponding short-horizon futures. RoboMeter \citep{robometer} scores the predicted outcomes, and the highest scoring candidate is selected for closed-loop execution. Both settings use the same $\pi_{0.5}$ policy and closed-loop execution protocol; the only difference is whether HF-WM prediction and RoboMeter scoring are used for test-time candidate selection.

\begin{minipage}{\linewidth}
    \centering
    \includegraphics[width=\linewidth]{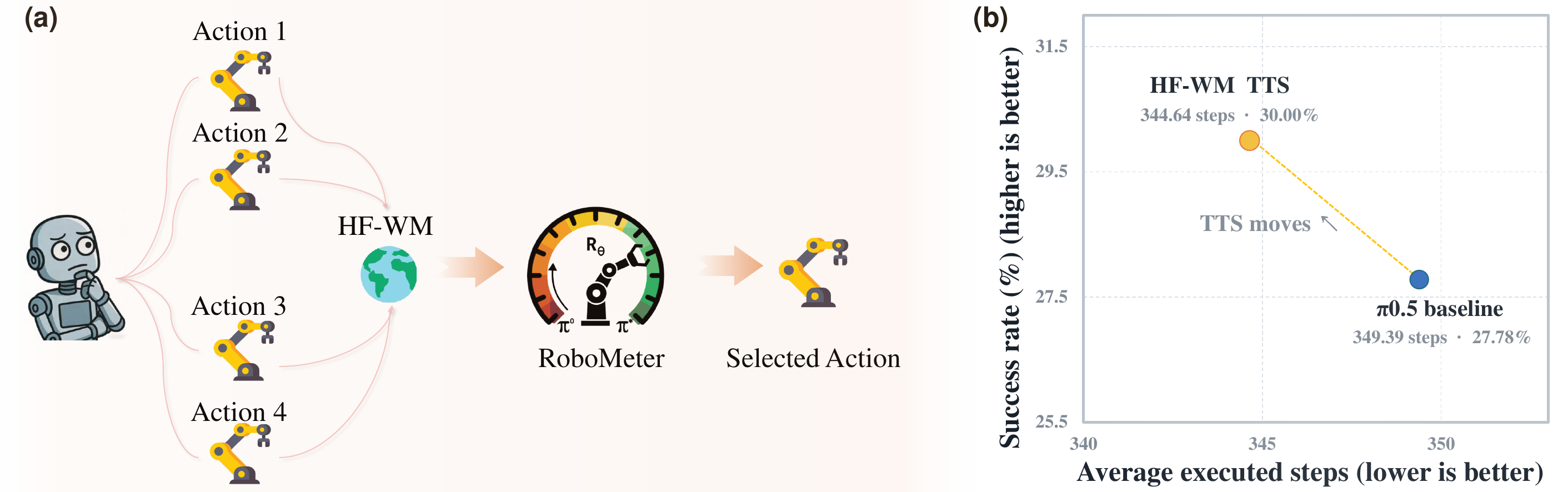}
    \captionof{figure}{Test-time scaling with HF-WM. (a) Candidate action selection with HF-WM and RoboMeter. (b) Improved success rate and reduced execution steps on LIBERO-90.}
    \label{fig:tts}
\end{minipage}

HF-WM-based test-time selection improves the closed-loop success rate from $27.8\%$ to $30.0\%$, while reducing the average episode length across the 90 tasks from $349.39$ to $344.64$ steps.
These results show that HF-WM enables RoboMeter to identify more promising action candidates at test time, translating policy sampling diversity into measurable closed-loop gains with more efficient execution and higher task success.

\begin{minipage}{\linewidth}
    \centering
    \includegraphics[width=\linewidth]{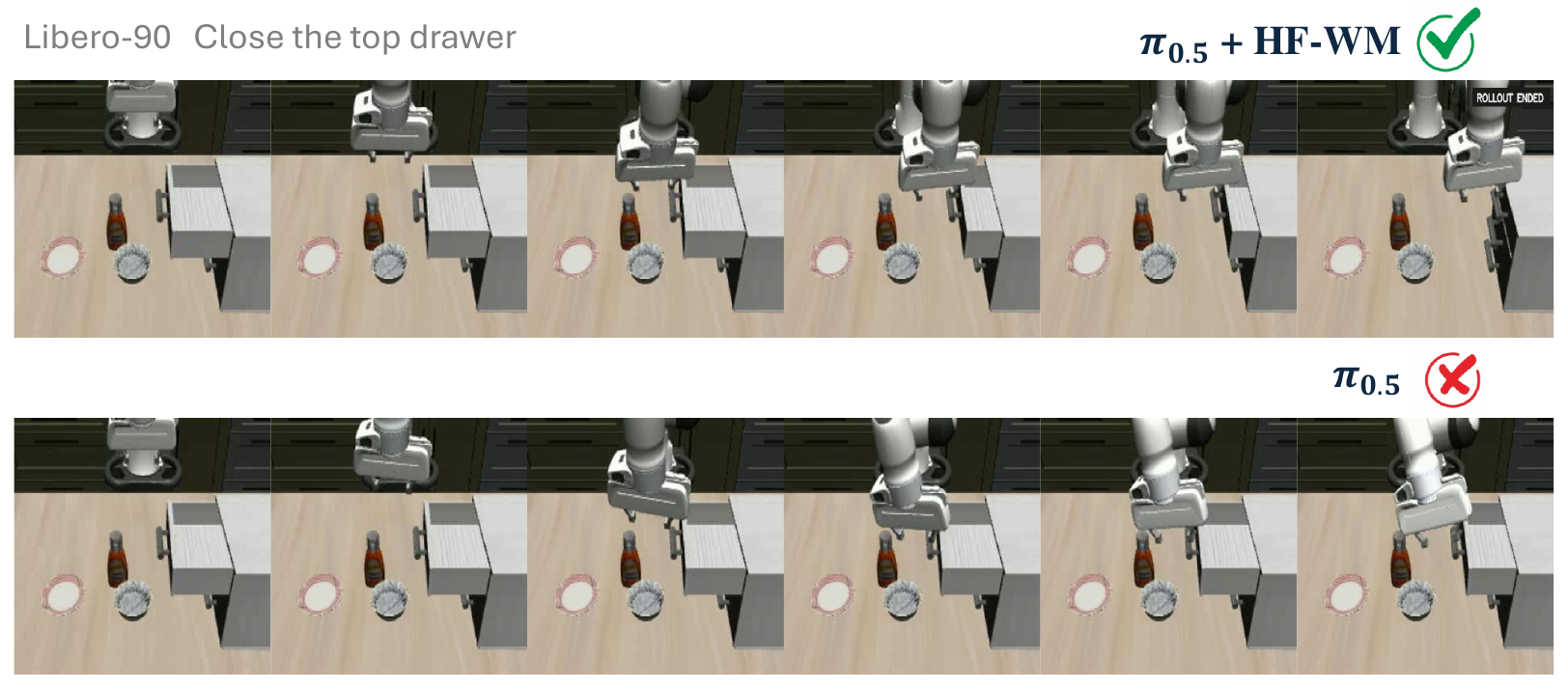}
    \captionof{figure}{An example episode from the LIBERO-90 benchmark. HF-WM with RoboMeter routing reduces action steps and improves success rate.
}
    \label{fig:libero}
\end{minipage}

\subsection{Additional Visualization Results}
\begin{figure}[!t]
    \centering
    \includegraphics[width=0.82\linewidth]{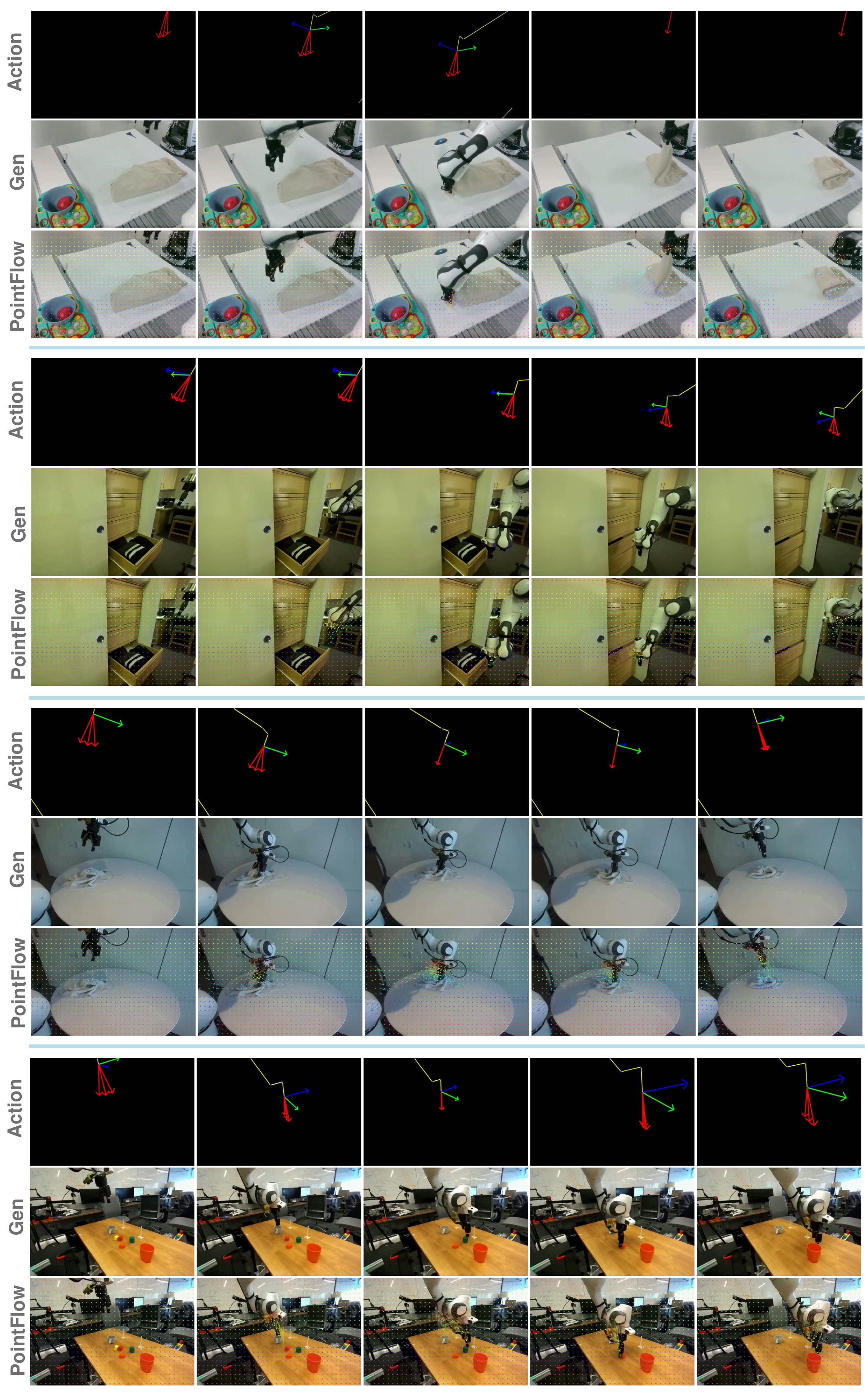}
    \caption{Additional qualitative results for HF-WM.}
    \label{fig:morecase}
\end{figure}

\begin{figure}[!t]
    \centering
    \vspace{-1cm}
    \includegraphics[width=0.82\textwidth]{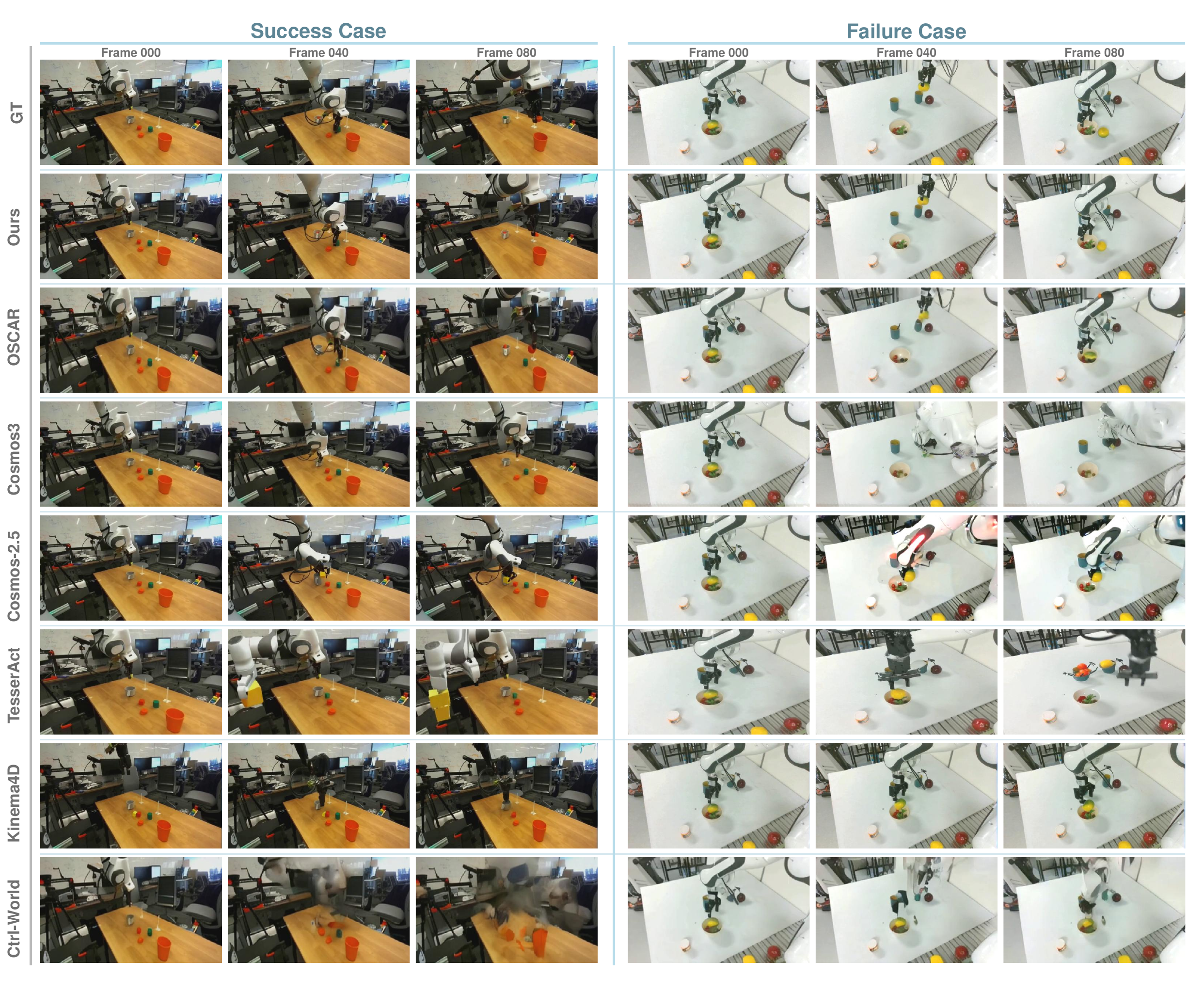}

    \vspace{-0.5cm}

    \includegraphics[width=0.82\textwidth]{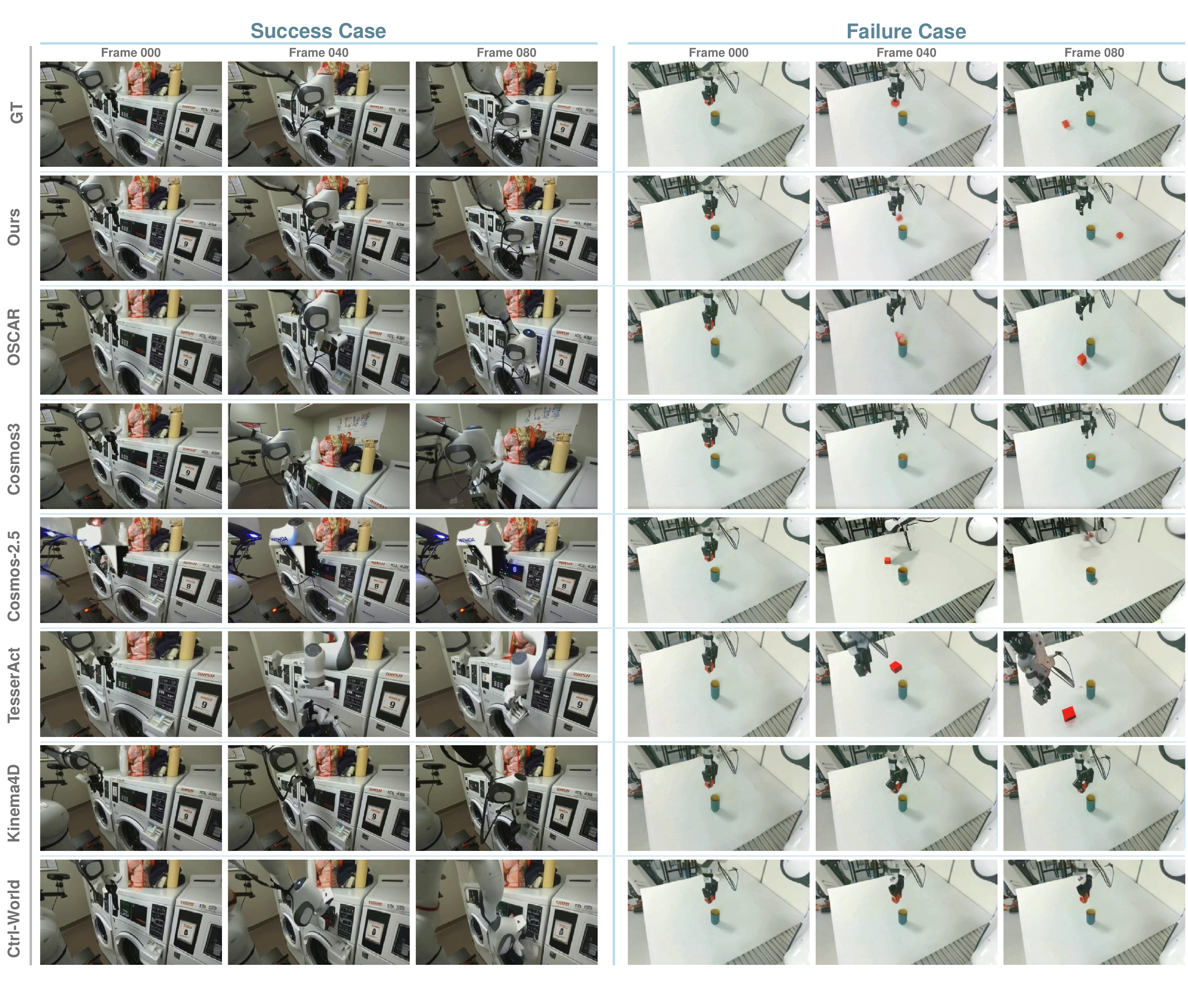}
    \vspace{-0.2cm}
    \caption{Additional qualitative comparison results}
    \label{fig:compare_more}
\end{figure}

We provide additional HF-WM rollouts with corresponding pointflow visualizations in Fig.~\ref{fig:morecase} and qualitative comparisons with representative baselines in Fig.~\ref{fig:compare_more}.

\clearpage

\end{document}